\documentclass{article}

\usepackage{microtype}
\usepackage{graphicx}
\usepackage{subcaption}
\usepackage{booktabs} 
\usepackage{wrapfig}
\usepackage{tabularx}

\usepackage{makecell}
\usepackage{xcolor}
\usepackage{amsmath}
\usepackage{hyperref}

\usepackage[accepted]{icml2026}

\usepackage{amsmath}
\usepackage{amssymb}
\usepackage{mathtools}
\usepackage{amsthm}
\usepackage{enumitem}

\usepackage[capitalize,noabbrev]{cleveref}

\theoremstyle{plain}

\theoremstyle{definition}

\theoremstyle{remark}

\usepackage[textsize=tiny]{todonotes}

\icmltitlerunning{Dive into the Scene: Breaking the Perceptual Bottleneck in Vision-Language Decision Making via Focus Plan Generation}

\begin{document}

\twocolumn[
  \icmltitle{Dive into the Scene: Breaking the Perceptual Bottleneck in Vision-Language \\ Decision Making via Focus Plan Generation}



  \icmlsetsymbol{equal}{*}

  \begin{icmlauthorlist}
    \icmlauthor{Boyuan Xiao}{zjucad}
    \icmlauthor{Bohong Chen}{zjucad}
    \icmlauthor{Yumeng Li}{zjucad}
    \icmlauthor{Ji Feng}{baiont}
    \icmlauthor{Yao-Xiang Ding}{zjucad,tz}
    \icmlauthor{Kun Zhou}{zjucad}
  \end{icmlauthorlist}

  \icmlaffiliation{zjucad}{State Key Lab of CAD\&CG, Zhejiang University, China}
  \icmlaffiliation{baiont}{Baiont Quant, China}
  \icmlaffiliation{tz}{Zhejiang Key Laboratory of Intelligent Medical Decision Support, China}
  \icmlcorrespondingauthor{Yao-Xiang Ding}{dingyx.gm@gmail.com}

  \icmlkeywords{vision-language model, decision-making, scene understanding}

  \vskip 0.3in
]



\printAffiliationsAndNotice{}  

\begin{abstract}
In embodied vision-language decision making tasks such as robotic manipulation and navigation, Vision-Language and Vision-Language-Action Models (VLMs \& VLAs) are powerful tools with different benefits: VLMs are better at long-term planning, while VLAs are better at reactive control. However, their performance is limited by the same perceptual bottleneck: visual hallucinations arise due to the models’ inability to distinguish task-relevant objects from distractors. In principle, accurate identification and focus on critical objects while filtering out irrelevant ones is the key to break this limitation. A straightforward solution is one-step focus: directly attending to essential objects. However, this approach proves ineffective because effective focus inherently requires deep scene understanding. To this end, we propose {\it SceneDiver}, a coarse-to-fine focus plan generation method for VLMs leveraging their long-term planning abilities, that first constructs a holistic scene graph to establish initial comprehension, then progressively decomposes the task into simpler sub-problems through an iterative cycle of recognition, understanding, and analysis. To enable reactive control, we also design a lightweight adapter for distilling the deliberate focus ability into VLAs.  Evaluations on standard embodied AI benchmarks confirm that our method substantially reduces visual hallucinations for both VLMs and VLAs, while preserving computational efficiency in tasks requiring fast execution. Our code and data are released at: \url{https://future-item.github.io/SceneDiver}.

\end{abstract}

\section{Introduction}
Embodied vision-language decision making tasks, such as robotic manipulation and navigation, unify the challenges of visual perception, language understanding, as well as planning and action~\citep{sapkota2025vision}. Vision-Language and Vision-Language-Action Models (VLMs \& VLAs) are powerful tools for embodied decision making with different benefits: VLMs are better as deliberate high-level planners~\citep{ahn2022icanisay,driess2023palmeembodiedmultimodallanguage}, whereas VLAs are better as end-to-end reactive executors~\cite{brohan2023rt2visionlanguageactionmodelstransfer, kim2025finetuningvisionlanguageactionmodelsoptimizing,black2026pi0visionlanguageactionflowmodel}. 

\begin{figure} [!t]
    \centering
    \includegraphics[width=\linewidth]{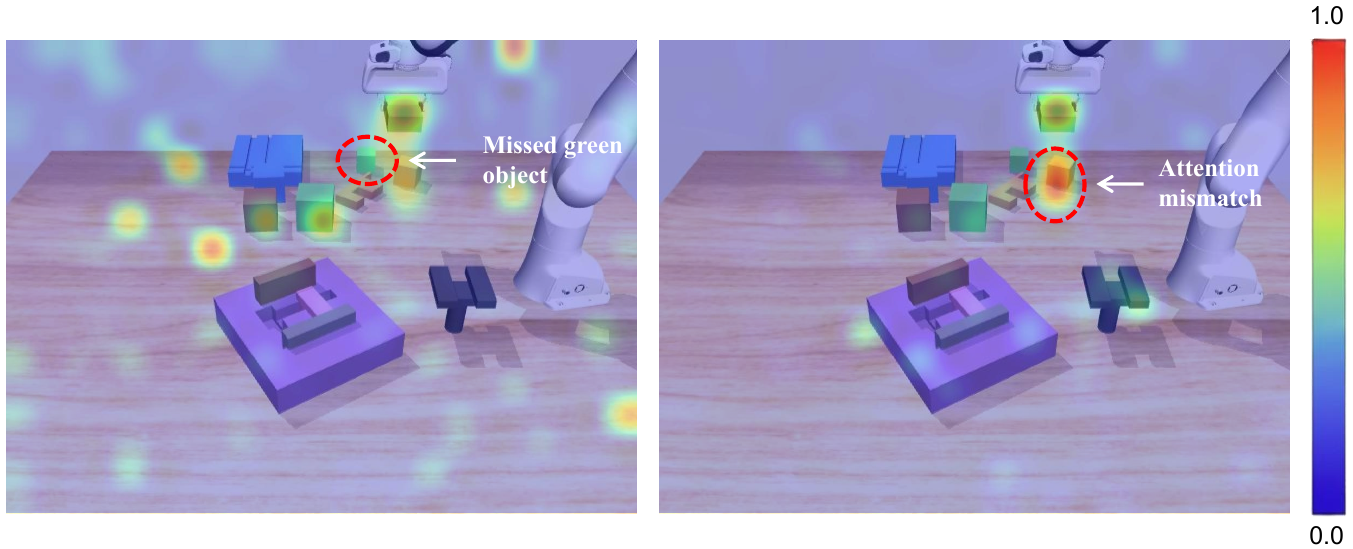}
    \caption{Failures in VLM decision making during direct text-image queries. Left: When queried about the number of green objects, the VLM's attention fails to capture the target objects, instead allocating excessive focus to irrelevant background elements. Right: When asked for the color of the object grasped by the robotic arm, the model exhibits attention mismatching; the focus incorrectly shifts to the yellow block rather than the target green object.}
    \label{fig:motivation} 
\end{figure}

However, despite their distinct functional advantages, both of them encounter a common perceptual bottleneck, manifested in object hallucination, where non-existent objects are spuriously predicted, and perceptual errors including object omission, erroneous attribute binding, and inaccurate instance counting within the same category. In principle, accurate identification and focus on critical objects while filtering out irrelevant ones is the key to break this limitation. A straightforward solution is one-step focus: directly attending to essential objects with existing established visual focusing strategies. However, decision making in complex visual environments often demands progressive scene exploration to gradually disentangle task-relevant objects from cluttered backgrounds and verify their identity across multiple scales, making one-step focus strategies ineffective. For example, as shown in Figure ~\ref{fig:motivation}, we present two failure cases of one-shot focus. It can be observed that without a comprehensive understanding of the scene, VLMs exhibit systematic mis-grounding and attention shifts.

To address this challenge, we propose {\it SceneDiver}, an effective method to break the perceptual limitation of VLMs in embodied vision-language decision making, leveraging their autonomous planning abilities. SceneDiver enables VLMs to autonomously generate the {\it focus plan}, which can help VLMs to focus only on task-relevant objects in each decision-making step. The central challenge lies in enabling VLMs to identify task-relevant objects from a cluttered scene containing numerous distractors. To address this, we introduce a coarse-to-fine, two-stage focus plan generation pipeline. At the first coarse-grained plan stage, we convert raw image data into structured graph representations by constructing scene graphs. Graph reasoning is then performed over the constructed graph, decomposing the complex global scene into a series of simpler local sub-scenes corresponding to individual nodes in the graph. At the second fine-grained plan stage, VLMs are requested to autonomously explore each local sub-scene in agentic manner to discover task-relevant objects. Finally, the visual input is refined using the objects identified and fed to the VLM for the final decision. Essential task-related information is preserved, while irrelevant content is suppressed. Furthermore, to meet the latency requirements of real-time decision making, we design a lightweight adapter that distills our progressive focus planning into VLA models, allowing them to benefit from breaking perceptual ability while maintaining inference efficiency suitable for online deployment.

We conduct experiments across diverse tasks of robotic manipulation~\citep{reflectVLM} and room navigation~\citep{yang2025embodiedbenchcomprehensivebenchmarkingmultimodal}. The results demonstrate that our method achieves a 10\%–15\% performance gain in manipulation tasks and up to a 16\% improvement in navigation, outperforming existing visual focus approaches. Then, we also evaluate our method on the LIBERO-plus benchmark~\citep{fei2025liberoplusindepthrobustnessanalysis} (which is employed to evaluate the robustness of the VLA against environmental interference). Experimental results demonstrate that our method achieves up to a 9.6\% improvement in success rate, enhancing decision-making robustness while incurring a marginal computational overhead of only 2.64\%.

\section{Related Work}
\subsection{Vision-Language Decision Making} 
Vision-language decision making has become a promising research area~\citep{ma2024survey}. Benefiting from the perceptual capabilities of MLLMs and their comprehension of natural language instructions, traditional control policies have become more diversified and are now capable of interacting with users in a more intuitive manner. For instance, ~\citep{brohan2023can} employs MLLMs as high-level task planners, enabling the decomposition of user-provided instructions into a sequence of plausible low-level skills. Similarly, ~\citep{huang2022language} adopts a two-stage framework to translate high-level instructions into executable actions.~\citep{jang2022bc} achieves zero-shot task generalization to unseen tasks by aligning language instructions or human demonstration videos with visual observations of the environment. The performance of decision making in such systems is closely tied to the effectiveness of the vision encoder. Accordingly, numerous approaches have been proposed to enhance the visual processing components of MLLMs.~\citep{shang2024theia} integrates knowledge distilled from diverse vision foundation models into a unified architecture to maximize visual representation capacity. ~\citep{assran2023self} constructs an implicit world model by comparing embeddings of image patches to capture structural regularities. ~\citep{karamcheti2023language} further improves vision-language alignment by incorporating language conditioning and generation into the masked image autoencoder (MAE) training objective.

\subsection{Object Hallucination}
Despite VLM's promising performance on various benchmarks~\citep{danish2025geobenchvlm,li2025survey,zong2024vlicl}, in complex visual scenes, these models~\citep{liu2023llava,chen2024spatialvlm} frequently perceive objects that do not exist in the provided images, a problem known as object hallucination~\citep{rohrbach2019objecthallucinationimagecaptioning,dai2023plausiblefaithfulprobingobject}. Prior work has explored several avenues to mitigate this issue, including integrating an external object detector~\citep{zhai2024hallecontrolcontrollingobjecthallucination}, applying visually grounded visual instruction tuning~\citep{you2023ferretrefergroundgranularity,zhang2024groundhoggroundinglargelanguage} or using reinforcement learning~\citep{sun2023aligninglargemultimodalmodels,gunjal2024detectingpreventinghallucinationslarge}.

While prior work focuses on general object hallucination, the specific challenge of multi-object hallucination—where models invent multiple entities with incorrect attributes and relationships—remains less explored. ~\citet{Multi-object-hallucination} finds that high cognitive loads from multi-object queries lead models to use heuristic shortcuts, bypassing rigorous visual analysis for each object.
To mitigate this, researchers explore several avenues. ~\citet{zhu2025mitigating} identifies that non-uniform spatial attention causes identical objects to be processed differently based on their position, and proposes a training-free attention rectification method to address this. Targeting a phenomenon termed ``local over-trust'', ~\citet{huang2024opera} develops a method that monitors attention weights during decoding, which penalizes generation paths where a single token excessively influences subsequent content.

\subsection{Scene Graphs}
Scene graphs have been widely studied in vision-language research. Early work formulated scene graphs as structured representations of objects, attributes, and relations for semantic image retrieval~\citep{johnson2015image}. Visual Genome~\citep{krishna2017visual} later established scene graphs as a large-scale resource for dense scene understanding by providing annotations of objects, attributes, and relationships. Scene graphs have subsequently been used in visual question answering and reasoning, for example in GQA~\citep{hudson2019gqa}, where they provide structured supervision and an explicit substrate for reasoning. Additionally, scene graphs have also been explored for image captioning, both as intermediate semantic representations for generation and as objects of analysis regarding their contribution to caption quality~\citep{yang2019auto,wang2019role}. Beyond vision-language tasks, scene-graph representations have also been extended to embodied and robotic settings. As representatives, Nguyen et al.~\citep{nguyen2025generating} leverages 3D scene graphs as an intermediate representation of robot environments, which are converted into knowledge graphs to enable actionable reasoning for robot decision-making. Terra~\citep{samuelson2025terra} constructs a hierarchical, terrain-aware 3D scene graph for task-agnostic outdoor mapping, underscoring the utility of structured scene representations in robotic reasoning. Finally, scene-graph representations have influenced evaluation as well: SPICE~\citep{anderson2016spice} compares candidate and reference captions through semantic tuples derived from scene-graph-like structures rather than relying solely on n-gram overlap. Different from existing approaches, we let the scene graphs play the role as a structured prior for perception, helping break the perceptual bottleneck by enabling coarse-to-fine focus planning over task-relevant objects in cluttered scenes.

\section{Method}
\subsection{Overview}
\begin{figure*}[!t]
    \centering  
    \includegraphics[width=0.9\linewidth]{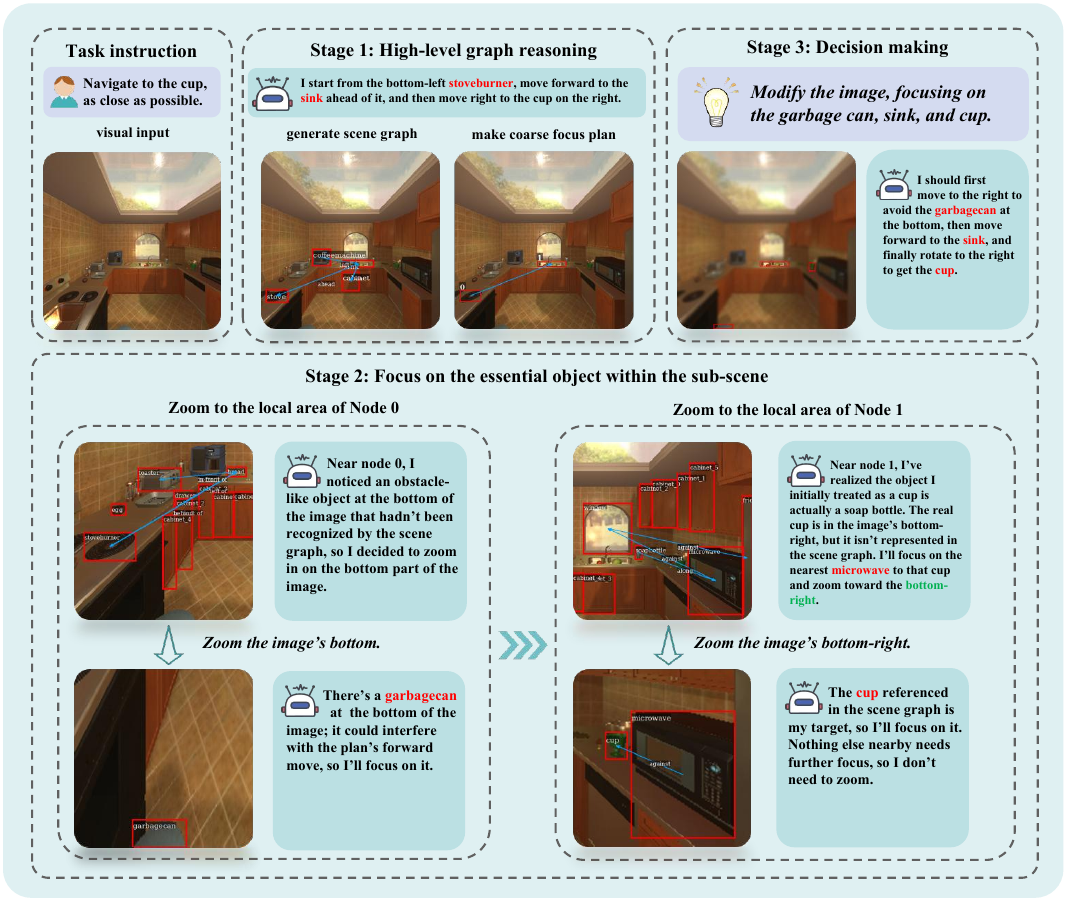} 
    \caption{Overview of SceneDiver: From the input image we build a scene graph, perform graph reasoning to decompose the complex global scene into a series of simpler local sub-scenes corresponding to individual nodes in the graph (Stage 1), autonomously explore each local sub-scene using naturally designed exploration strategies to identify task-relevant objects (Stage 2), and use the resulting focus to modify the image and make a decision.} 
    \label{fig:pipeline}   
\end{figure*}
Vision-Language Models (VLMs) frequently struggle with visual hallucinations and attention drift. While holistic scene understanding is essential and naive filtering risks discarding necessary information, processing the full, uncurated view exposes the model to task-irrelevant background clutter and spurious features that can mislead decision making. To address this, we introduce SceneDiver, a focus
plan generation method that guides the model’s attention through a coarse-to-fine reasoning process. In essence, we synthesize holistic graph-based reasoning with modulated perception to enhance decision making. Specifically, our system first executes high-level reasoning using a scene graph to identify critical regions of interest, denoted as \emph{sub-scenes}. It then ``looks closely'' at these sub-scenes using a restricted perception field to iteratively verify their relevance. This process yields a pixel-level focus map that visually suppresses distracting content, effectively steering the model's attention toward essential targets. Finally, to facilitate real-time reactive control, we distill this explicit planning capability into a lightweight adapter for Vision-Language-Action (VLA) models.

\subsection{Coarse Stage: Graph Reasoning}
We utilize scene graphs as a structured prior to provide a coarse guide for the VLM's reasoning process. Using an OvSGTR~\cite{chen2024expandingscenegraphboundaries} backend, we generate a graph representation where nodes denote objects and edges encode spatial relationships. We input the scene graph information into the VLM in textual format. Specifically, detected objects are annotated with \texttt{<ref>}, and their spatial relationships are specified by \texttt{<pred>}. The location of each object is encoded using \texttt{<box>}. On this structure, the VLM executes high-level graph reasoning to identify task-relevant sub-scenes for subsequent processing. During this process, we require the VLM to produce concise, structured outputs at each step, using the graph's unique identifiers (\texttt{<ref>}) and spatial coordinates (\texttt{<box>}) to mark its regions of interest. We treat the VLM as the single source of truth and the graph as a reasoning guide for traversal. When inconsistencies arise, the VLM is free to select spatially proximate neighbors, discard if a target is irrelevant, or to retain ambiguous sub-scene as provisional candidates for the next stage.

\subsection{Fine Stage: Verification and Exploration}
The coarse graph reasoning provides a high-level route of sub-scenes, which is followed by a stage where the VLM sequentially zooms in on and processes each sub-scene. The model actively verifies whether information is relevant to the current task using a limited perception field, and autonomously selects, discards, or searches nearby areas to locate relevant information in an agentic manner. If the object within the local window aligns with the plan, it is immediately confirmed as a valid candidate, and the process moves to the next sub-scene. If the visual evidence is ambiguous, the model triggers a \emph{semantic zoom}, narrowing its field of view to resolve fine-grained details. Conversely, if the expected target is absent due to graph-to-image misalignment, the model initiates a \emph{local spatial search}, scanning the immediate neighborhood to attempt to re-acquire the target. This agentic behavior allows the system to recover from resolution limits or spatial drift without requiring hard-coded rules, naturally building the final set of verified objects $\mathcal{C}$ by solely utilizing the inherent capabilities of the VLM.

\subsection{Guiding Decision through Focus Modulation}
Upon establishing the candidate set $\mathcal{C}$ from the fine stage, we rasterize the candidate bounding boxes $\{b_k\}_{k\in\mathcal{C}}$ into a pixel-level Focus Score Map $s \in [0,1]^{H \times W}$, defined as $s_{u,v}=\mathbb{I}\!\left[\exists\,k\in\mathcal{C}:\ (u,v)\in b_k\right]$. To direct attention without discarding environmental context, we empirically design a soft modulation that simultaneously attenuates luminance and high-frequency details in background regions based on the map. Given the input image $I$ and a visibility floor $\beta$ (to prevent total blackout), we first compute a luminance-scaled version $I_{\text{dim}}$. We then synthesize the final composite $I_{\text{out}}$ by blending this scaled image with its Gaussian-blurred counterpart $\mathcal{B}_{\sigma}$ inversely proportional to the score:
\begin{align}
    I_{\text{dim}} &= I \odot \bigl(\beta + (1-\beta)s\bigr), \\
    I_{\text{out}} &= s \odot I_{\text{dim}} \;+\; (1-s) \odot \mathcal{B}_{\sigma}(I_{\text{dim}}).
\end{align}
This coupled operation keeps target objects sharp and bright while clutter is gently dimmed to reduce visual interference, fusing score map and raw image prompt into a modulated image prompt.

\subsection{SceneDiver Adapter}
\begin{figure*}[!htbp]
    \centering  
    \includegraphics[width=0.9\linewidth]{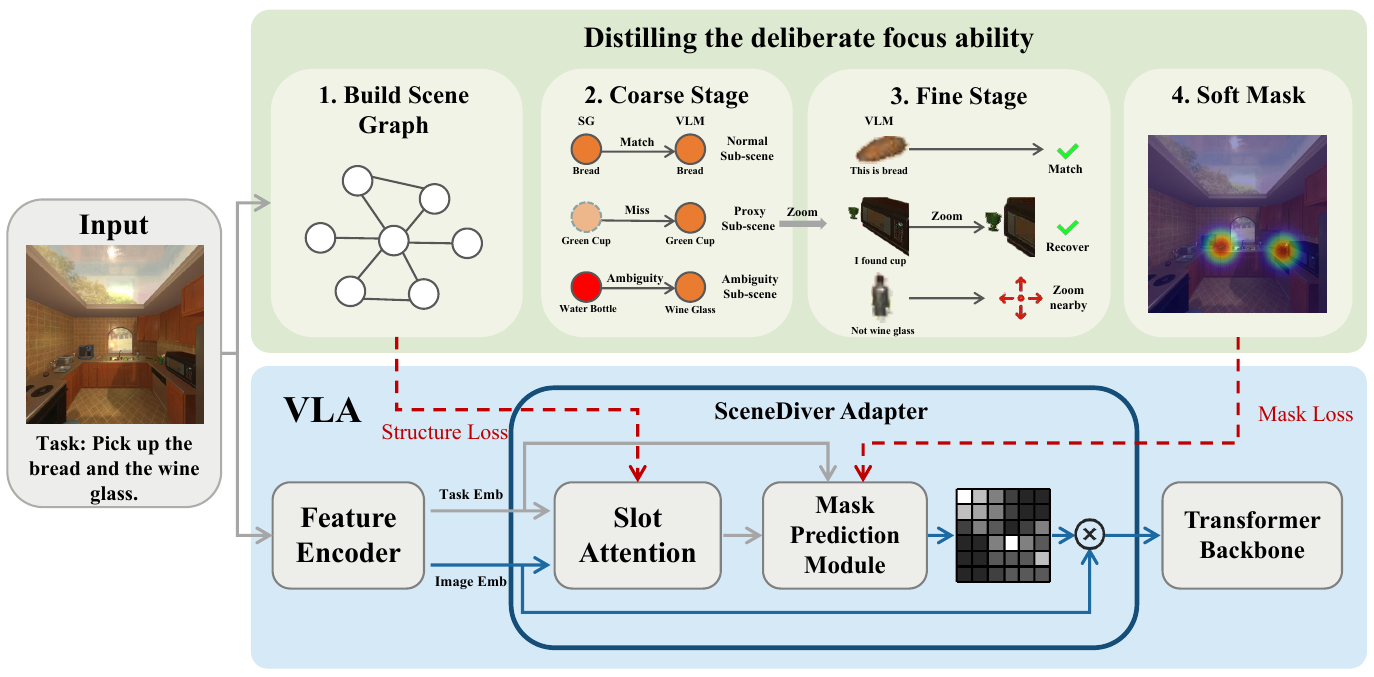} 
    \caption{Distilling SceneDiver into a lightweight adapter to transfer the deliberate focus ability to a VLA for reactive control. Slot Attention learns the structured representations of scene graphs, while the mask prediction module learns the two-stage reasoning process.} 
    \label{fig:distillation}   
\end{figure*}
Iterative scene graph traversal enables the VLM to make better decisions, while remaining computationally intensive. We further construct a real-time solution which we denote as SceneDiver adapter below, an end-to-end VLA adapter that distills explicit reasoning into implicit representations. Positioned after the projector in the cross-modal alignment space, it employs Slot Attention~\cite{locatello2020objectcentriclearningslotattention} to extract task-conditioned structural representations, bridging visual tokens and logical entities. A mask prediction module then generates refined attention masks for task-relevant objects. Knowledge transfer is realized via two distillation objectives: a \emph{Structure Loss} supervising Slot Attention to learn structured reasoning-related representations, and a \emph{Mask Loss} supervising the mask prediction module to improve spatial localization accuracy.

\paragraph{Task-Guided Slot Attention.} We project visual features $F \in \mathbb{R}^{L \times D}$ into object-centric slots $S \in \mathbb{R}^{K \times D_{s}}$, where \(L\) denotes the number of visual patches, \(K\) denotes the number of object-centric slots, and \(D\) denotes the feature dimension of each patch (with \(D_s\) the slot feature dimension). Unlike standard random initialization, we initialize this network by the following way: task tokens $T$ are pooled via learned attention into a global vector $v_{task}$. This vector conditions the initialization mean while variance remains a learnable global parameter:
\begin{equation}
    S_{init} \sim \mathcal{N}(\mu(v_{task}) + \delta, \sigma_{global}).
\end{equation}
This encourages slots to capture task-relevant entities rather than arbitrary textures.

\paragraph{Scene-Aware Mask Prediction Module.} This module performs hierarchical reasoning analogous to the coarse-to-fine mechanism. At the \textbf{coarse stage}, it fuses slot semantics, slot mass (attention weight reflecting object scale), and task context to predict relevance scores $r_k$. At the \textbf{fine stage}, existing attention maps $A \in \mathbb{R}^{K \times L}$ propagate slot semantics back to patches, producing residual corrections:
\begin{equation}
    M_{pred} = \sigma\left( \sum_k r_k \cdot A_{k,:} + \alpha \cdot \Delta_{patch} \right),
\end{equation}
where $\alpha$ is initialized near zero, allowing the network to rely initially on slot-level predictions before progressively incorporating spatial refinements; $\Delta_{patch}$ is a per-patch residual correction predicted at the fine stage, $\sigma(\cdot)$ is the sigmoid function, and $M_{pred}$ is the predicted patch-level mask probability.

\paragraph{Distillation and Safety.} We train the adapter via Hungarian Matching~\citep{kuhn1955hungarian} between slots and scene graph-derived objects, targeting two levels: slot-level alignment for correct scene decomposition, and mask-level refinement for spatially precise focus maps. For robust deployment, we introduce entropy-based dynamic gating. When uncertainty (measured over ambiguous patches) exceeds a threshold, the system bypasses masking and feeds raw observations to the VLA, allowing for graceful degradation in ambiguous scenarios.

\section{Experiments}
\begin{figure*}[!htbp]
    \centering  
    \includegraphics[width=0.95\textwidth]{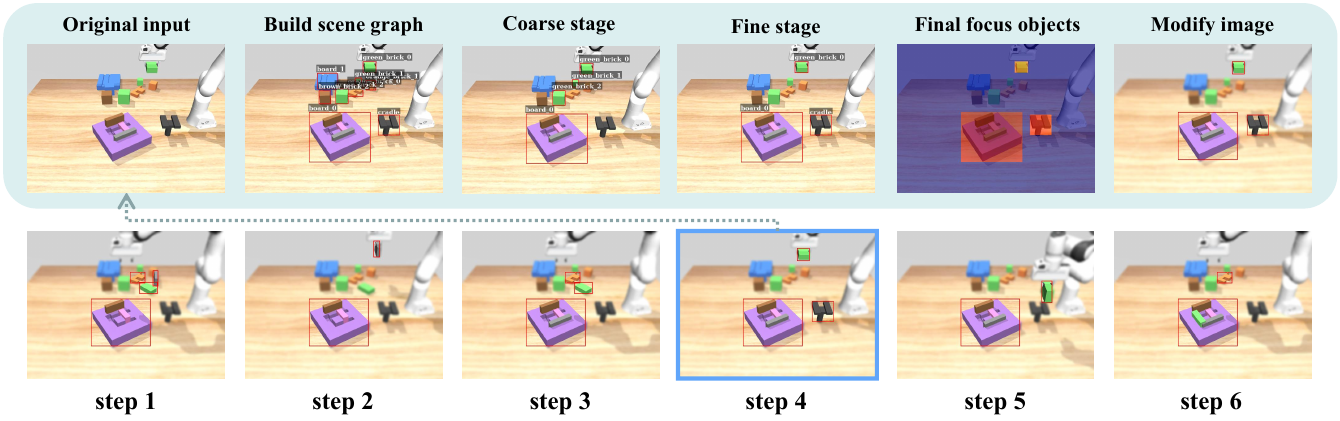} 
    \caption{Qualitative visualization of the SceneDiver execution process for robot manipulation. The focus of each step is highlighted with a red bounding box (for visualization purposes only; the box is invisible to the VLM). The workflow of SceneDiver is illustrated using Step 4 as a representative example.} 
    \label{fig:robot}   
\end{figure*}
Our experimental evaluation is organized into four main phases. First, we present the qualitative operation of SceneDiver and compare its performance against baselines in the context of robotic manipulation~\cite{reflectVLM}. Second, we evaluate our method on room navigation tasks~\citep{yang2025embodiedbenchcomprehensivebenchmarkingmultimodal}, benchmarking it against existing focus approaches designed to mitigate visual hallucinations; Experimental results demonstrate that our method significantly outperforms other focus approaches in VLM-based decision making tasks. Third, we evaluate our method on LIBERO-Plus, a benchmark specifically designed for VLA robustness evaluation, using OpenVLA-OFT as the base model. Experimental results indicate that our method effectively bolsters the robustness of the base model while incurring minimal computational overhead. Finally, we perform comprehensive ablation studies to validate the necessity of each algorithmic component and demonstrate the system's robustness against noisy or erroneous scene graphs through stress testing.

\subsection{Robotic Manipulation} 
In the first experiment, we constructed a robotic‐arm task in MuJoCo~\citep{todorov2012mujoco} following the setup of ~\citet{reflectVLM}. The scene consists of a base plate with small pieces randomly scattered across a tabletop, augmented by a decoy board and several decoy pieces that serve as distractors. The agent's objective is to distinguish the target bricks from the distractors and assemble them onto the base plate. The control policy utilizes a discrete action space: {\it pick up}, {\it insert}, {\it reorient}, and {\it put down}. We randomly sampled 30 distinct scenes to formulate challenging experimental instances. In each instance, the VLM is provided with images of the current and target states and must complete the assembly within a strict horizon of 30 steps. 
The results are reported in Table~\ref{tab:robotplan}, showing that our method generalizes effectively across multiple backends and significantly enhances decision accuracy in cluttered settings.

\begin{table}[!htbp]
  \centering
  \small
  \caption{Robot manipulation accuracy(Success Rate \%). Means and Standard Errors (in the parentheses) over 5 seeds are reported.}
  \renewcommand{\tabularxcolumn}[1]{m{#1}}
  \begin{tabularx}{.8\linewidth}{>{\raggedright\arraybackslash}X c c} 
    \toprule
    \textbf{Model}       & \textbf{Base} & \textbf{Focus} \\ 
    \midrule
    Qwen2.5-VL-7B-Instruct-AWQ ~\citep{Qwen2.5-VL}      & $\text{14.7}_{(\text{1.8})}$    & \boldmath{$\text{28.7}_{(\text{1.6})}$}     \\ 
    \midrule
    Qwen2.5-VL-32B-Instruct-AWQ ~\citep{Qwen2.5-VL}     & $\text{21.3}_{(\text{1.2})}$    & \boldmath{$\text{31.3}_{(\text{0.8})}$}     \\ 
    \midrule
    gpt-4o-mini~\citep{openai2024gpt4o}            & $\text{28.7}_{(\text{1.6})}$    & \boldmath{$\text{34.0}_{(\text{1.2})}$}    \\ 
    \midrule
    gemini-2.5-flash~\citep{comanici2025gemini25pushingfrontier}        & $\text{38.7}_{(\text{1.2})}$  &  \boldmath{ $\text{46.7}_{(\text{1.0})}$}    \\ 
    \midrule
  \end{tabularx}
  \label{tab:robotplan} 
\end{table}

We illustrate a concrete decision trajectory in Figure ~\ref{fig:robot}. The red boxes are for visualization only and are not visible to the VLM. We present the process of SceneDiver at step 4. In the coarse stage, the VLM infers from the task and scene graph reasoning that the green block on the robotic arm may be the assembly target. However, since there are multiple green objects in the scene, it cannot determine whether the grasped object is the correct one. Therefore, in the fine stage, the VLM focuses on each green block individually, discovering that while the block on the gripper has the correct shape, its orientation is incorrect and requires rotation. Consequently, attention must also be paid to the cradle for reorienting the block. Ultimately, the VLM identifies that attention should be directed to the green block on the robotic arm, the board, and the cradle.

\subsection{Room Navigation}

\newcommand{\NA}{-}         
\newcommand{\NApm}{-\,(-)}  
\begin{table*}[!htbp]
\centering
\caption{Quantitative comparisons (Success Rate \%) over 5 random seeds under the room navigation benchmark. \textbf{CS}: Common Sense, \textbf{CI}: Complex Instruction, \textbf{VA}: Visual Appearance.}
\label{tab:navigation}
\small

\newcolumntype{Y}{>{\centering\arraybackslash}X}

\setlength{\tabcolsep}{1pt} 

\newcommand{\res}[2]{#1\textcolor{gray}{\tiny (#2)}}

\begin{tabularx}{\linewidth}{@{} l | *{12}{Y} @{}} 
\toprule
& \multicolumn{12}{c}{\textbf{Open-Source Models}} \\
\cmidrule(lr){2-13}
& \multicolumn{4}{c|}{\makecell{Qwen2.5-VL-7B-Instruct-AWQ \\ \scriptsize \citep{Qwen2.5-VL}}} & \multicolumn{4}{c|}{\makecell{Qwen2.5-VL-32B-Instruct-AWQ \\ \scriptsize \citep{Qwen2.5-VL}}} & \multicolumn{4}{c}{\makecell{InternVL2.5-8B \\ \scriptsize \citep{chen2024internvl}}} \\
Method & Base & CS & CI & VA & Base & CS & CI & VA & Base & CS & CI & VA \\
\midrule

Base Model    & \res{32.7}{1.5} & \res{30.7}{0.6} & \res{32.0}{1.2} & \res{27.3}{2.4} & \res{49.3}{1.1} & \res{43.3}{1.9} & \res{46.7}{1.3} & \res{43.3}{0.9} & \res{29.3}{1.1} & \res{22.7}{1.1} & \res{24.0}{1.1} & \res{21.3}{1.5} \\
SoM           & \res{30.0}{0.9} & \res{31.3}{1.2} & \res{31.3}{1.8} & \res{29.3}{2.2} & \res{49.3}{1.1} & \res{44.7}{1.2} & \res{48.0}{0.7} & \res{44.7}{1.5} & \res{28.0}{1.5} & \res{24.0}{1.1} & \res{23.3}{0.9} & \res{24.0}{1.1} \\
Multi-Res     & \res{29.3}{1.7} & \res{32.7}{0.6} & \res{34.0}{1.1} & \res{29.3}{1.1} & \res{52.7}{1.1} & \res{46.7}{1.6} & \res{49.3}{1.1} & \res{45.3}{2.0} & \res{31.3}{2.8} & \res{20.7}{0.6} & \res{30.7}{1.5} & \res{22.7}{1.1} \\
VCD           & \res{34.7}{2.0} & \res{32.0}{2.2} & \res{32.7}{2.4} & \res{33.3}{2.1} & \res{51.3}{1.5} & \res{46.0}{1.5} & \res{45.3}{1.5} & \res{46.0}{1.1} & \res{33.3}{1.3} & \res{24.7}{2.0} & \res{28.0}{0.7} & \res{23.3}{0.9} \\
\midrule

\textbf{Ours} & \textbf{\res{44.0}{1.1}} & \textbf{\res{36.0}{0.6}} & \textbf{\res{37.3}{0.6}} & \textbf{\res{35.3}{0.7}} & \textbf{\res{56.7}{1.3}} & \textbf{\res{52.7}{1.1}} & \textbf{\res{53.3}{0.9}} & \textbf{\res{49.3}{1.1}} & \textbf{\res{39.3}{0.7}} & \textbf{\res{32.7}{0.6}} & \textbf{\res{33.3}{1.1}} & \textbf{\res{25.3}{0.7}} \\
\bottomrule
\end{tabularx}

\vspace{1.2em}

\begin{tabularx}{\linewidth}{@{} l | *{12}{Y} @{}}
\toprule
& \multicolumn{12}{c}{\textbf{Closed-Source Models}} \\
\cmidrule(lr){2-13}
& \multicolumn{4}{c|}{\makecell{gpt-4o-mini \\ \scriptsize \citep{openai2024gpt4o}}} & \multicolumn{4}{c|}{\makecell{gemini-2.5-flash \\ \scriptsize \citep{comanici2025gemini25pushingfrontier}}} & \multicolumn{4}{c}{\makecell{doubao-seed-1.6-flash \\ \scriptsize \citep{guo2025seed15vltechnicalreport}}} \\
Method & Base & CS & CI & VA & Base & CS & CI & VA & Base & CS & CI & VA \\
\midrule

Base Model    & \res{43.3}{1.6} & \res{37.3}{2.4} & \res{38.7}{1.5} & \res{37.3}{1.7} & \res{68.0}{1.1} & \res{62.0}{0.7} & \res{60.0}{0.9} & \res{55.3}{1.2} & \res{59.3}{1.1} & \res{40.0}{1.3} & \res{38.0}{2.2} & \res{32.7}{1.5} \\
SoM           & \res{41.3}{0.7} & \res{38.0}{2.0} & \res{36.0}{1.1} & \res{38.7}{1.5} & \res{69.3}{1.1} & \res{62.7}{1.7} & \res{60.7}{2.4} & \res{56.0}{2.6} & \res{60.7}{2.6} & \res{40.0}{1.6} & \res{40.7}{1.1} & \res{34.0}{0.6} \\
Multi-Res     & \res{44.7}{1.5} & \res{38.7}{1.2} & \res{41.3}{2.0} & \res{43.3}{1.9} & \res{70.0}{1.9} & \res{60.7}{1.5} & \res{61.3}{2.0} & \res{58.0}{2.9} & \res{62.0}{2.4} & \res{41.3}{2.0} & \res{42.0}{1.2} & \res{37.3}{2.4} \\
Thinking      & - & - & - & - &     \res{70.0}{1.3} & \res{63.3}{1.3} & \res{63.3}{1.9} & \res{56.7}{1.0} &            \res{60.7}{2.2} &  \res{41.3}{1.2} & \res{43.3}{1.0} & \res{36.7}{1.3} \\
\midrule

\textbf{Ours} & \textbf{\res{50.0}{0.9}} & \textbf{\res{53.3}{0.9}} & \textbf{\res{49.3}{1.1}} & \textbf{\res{49.3}{0.6}} & \textbf{\res{74.7}{1.2}} & \textbf{\res{65.3}{0.7}} & \textbf{\res{66.0}{0.6}} & \textbf{\res{62.0}{1.2}} & \textbf{\res{66.0}{1.1}} & \textbf{\res{48.0}{0.7}} & \textbf{\res{48.7}{1.2}} & \textbf{\res{44.7}{0.7}} \\
\bottomrule
\end{tabularx}
\end{table*}

Building upon the room-navigation benchmark by~\citet{yang2025embodiedbenchcomprehensivebenchmarkingmultimodal}, we propose a more demanding variant designed to push the limits of decision making. We retain the original protocol where the agent navigates using only visual observations and environmental feedback, but introduce two key challenges. First, we select scenes with substantially higher environmental complexity. Second, we make target objects harder to identify by making them smaller, more frequently occluded, and easily confusable with visually similar distractors, further increasing perceptual ambiguity.

We evaluate our method across four distinct sub-tasks to test different capabilities: (1) Base: uses concise, direct instructions; (2) Common-Sense: presents instructions in natural, conversational language; (3) Complex-Instruction: requires identifying targets within lengthy, intricate prompts; and (4) Visual-Appearance: specifies targets indirectly via shape and color attributes rather than semantic names. We benchmark both open-source and closed-source models, conducting five independent trials per task. Table~\ref{tab:navigation} reports the mean decision accuracy and standard error these runs.

As detailed in Table \ref{tab:navigation}, we benchmark our approach against four representative paradigms. First, regarding visual prompting methods like SoM \cite{yang2023setofmarkpromptingunleashesextraordinary}, while its SAM-based tagging effectively discretizes objects, the dense markers often introduce excessive visual noise in cluttered scenes, obstructing the VLM's reasoning; conversely, our two-stage filtering isolates relevant objects without occlusion. Second, following the multi-scale input paradigm established by Monkey~\cite{li2024monkeyimageresolutiontext}, LLaVA~\citep{li2024llavaonevisioneasyvisualtask}, and Sphinx~\citep{lin2023sphinxjointmixingweights}, we feed images at varying resolutions (global low-res and local high-res crops) into VLMs. This method enhances fine-grained details but loses spatial information; in contrast, our method leverages Scene Graphs to preserve spatial topology and restricts zooming to selective candidates to ensure efficiency. Third, distinct from decoding strategies like VCD \cite{leng2023mitigatingobjecthallucinationslarge} which mitigate language priors by modifying output logits but fail to address root perceptual insufficiencies, our method actively enhances input fidelity, ensuring decisions are grounded in high-resolution visual evidence rather than statistical corrections. Finally, even thinking-enhanced models such as doubao-seed-1.6-flash-thinking and gemini-2.5-flash-thinking underperform our method, underscoring that extended reasoning alone cannot remedy perceptual deficiencies.

\subsection{End-to-End Control: Robustness on LIBERO-plus}
\newcommand{\res}[5]{%
    \makecell[c]{%
        \textbf{#3}\tiny$\pm$#4\% {\color{teal}\scriptsize($\uparrow$#5)} \\[-0.5ex] 
        {\color{gray}\scriptsize #1\tiny$\pm$#2\%} 
    }%
}

\begin{table*}[!htbp]
    \centering
    \small
    \caption{Performance comparison on LIBERO-plus benchmark suites. We report the mean success rate and standard error over 3 seeds.}
    \label{tab:libero_results}
    \renewcommand{\arraystretch}{1.4} 
    \setlength{\tabcolsep}{3pt}       
    \begin{tabular}{lccccc}
        \toprule
        \textbf{Task Suite} & \textbf{Objects Layout} & \textbf{Camera Viewpoints} & \textbf{Robot Initial States} & \textbf{Background Textures} & \textbf{Light Conditions} \\
        \midrule
        
        Spatial & \res{92.39}{0.27}{94.43}{0.41}{2.04} & \res{49.16}{0.56}{54.35}{0.99}{5.19} & \res{19.91}{0.74}{29.49}{0.30}{9.58} & \res{82.67}{0.79}{87.41}{0.40}{4.74} & \res{93.86}{0.18}{97.37}{0.88}{3.51} \\
        \midrule
        
        Object & \res{76.01}{0.27}{76.82}{0.27}{0.81} & \res{62.50}{0.51}{64.56}{0.86}{2.06} & \res{14.36}{0.27}{15.72}{0.00}{1.36} & \res{89.74}{0.66}{90.61}{1.53}{0.87} & \res{98.99}{0.67}{99.50}{0.17}{0.51} \\
        \midrule
        
        Goal & \res{49.36}{0.25}{52.53}{0.38}{3.17} & \res{51.02}{0.43}{52.74}{0.72}{1.72} & \res{12.56}{0.26}{18.85}{0.00}{6.29} & \res{81.78}{0.40}{91.09}{1.21}{9.31} & \res{93.94}{0.76}{97.35}{0.76}{3.41} \\
        \midrule
        
        Long & \res{65.28}{0.76}{74.52}{0.56}{9.24} & \res{40.13}{0.40}{42.97}{0.54}{2.84} & \res{32.32}{0.40}{37.73}{0.00}{5.41} & \res{81.07}{0.00}{90.00}{0.00}{8.93} & \res{84.47}{0.76}{92.42}{0.00}{7.95} \\
        
        \bottomrule
    \end{tabular}
\end{table*}
Next, we evaluate the effectiveness of our adapter on real-time decision-making VLA. Specifically, we validate  whether our proposed adapter can effectively distill the deliberate focus ability into the policy, thereby improving its resilience against environmental interference compared to the OpenVLA-OFT baseline. We conduct these evaluations on the challenging LIBERO-plus benchmark. Unlike standard benchmarks, LIBERO-plus serves as a rigorous stress test for policy robustness, introducing complex lighting variations, texture randomization, and adversarial distractors to the standard LIBERO suite.

We evaluate in five interference types: \textit{Objects Layout}, which involves randomized object placements; \textit{Camera Viewpoints}, which varies camera perspectives; \textit{Robot Initial States}, which randomizes the initial configurations of the robotic arm; \textit{Background Textures}, which applies randomized background textures; and \textit{Light Conditions}, which introduces randomized lighting variations.

Following the LIBERO-plus evaluation protocol, we trained both OpenVLA-OFT and our method exclusively on the original, interference-free LIBERO dataset. We employed a joint training strategy using a data mixture from the \textit{libero spatial}, \textit{libero object}, \textit{libero goal}, and \textit{libero long} suites, and subsequently evaluated performance on each task individually. The quantitative results are presented in Table \ref{tab:libero_results}, revealing a consistent performance advantage of our SceneDiver-augmented policy over the OpenVLA-OFT baseline across all task suites. Most notably, the proposed method exhibits exceptional resilience in the Long-horizon task suite, yielding substantial improvements of 9.2\% in Objects Layout and 8.0\% in Light Conditions. Since long-horizon manipulation is particularly susceptible to error accumulation triggered by visual ambiguity, these significant margins indicate that our latent instantiation effectively stabilizes the policy's attention, preventing the "attention drift" often observed in the baseline when environmental variables fluctuate.

Beyond temporal stability, the method demonstrates robust generalization against visual domain shifts. Across the Spatial, Object, and Goal suites, we observe steady gains, up to 9.6\%. This suggests that by explicitly modeling the scene structure in the latent space, our approach successfully disentangles semantic object information from irrelevant high-frequency pixel variations. Consequently, the policy maintains robust execution even when the visual appearance deviates significantly from the training distribution, effectively mitigating the structural and semantic ambiguities highlighted in our motivation.

\subsection{Ablation and Stress testing}
In this section, we conduct ablation studies and scene graph stress tests on the room navigation task. 

\textbf{Ablation 1: Efficacy of Textual Scene Graphs.} To assess whether structured scene knowledge alone suffices, we supplied scene graphs directly as auxiliary text input for decision making. Results indicate that while scene graphs provide useful priors, excessive textual verbosity disperses attention from visual features and introduces noise, yielding only marginal improvements. 

\textbf{Ablation 2: Step-by-Step vs. Direct Focusing.} To validate the necessity of our two-stage planning strategy, we compare it against a "Direct Focus" baseline that directly predicts targets given scene graphs without coarse-to-fine reasoning. Results indicate that this approach only focuses on the target object itself without considering the surrounding scene context, making it prone to confusion when similar-looking distractors are present. 

\textbf{Ablation 3: Impact of Fine-Grained Granularity.} Complementing our global planning, our method enforces fine-grained focus at each node. To assess the criticality of this component, we evaluated performance using only coarse-grained planning without fine-grained verification. Results reveal that coarse planning alone introduces hallucinated objects due to imprecise grounding, confirming that fine-grained verification is a critical correction mechanism.

\begin{table*}[!htbp]
\centering
\caption{Ablation results (Success Rate \%) over 5 random seeds under the room navigation benchmark. \textbf{CS}: Common Sense, \textbf{CI}: Complex Instruction, \textbf{VA}: Visual Appearance.}
\label{tab:ablation}
\small

\newcolumntype{Y}{>{\centering\arraybackslash}X}

\setlength{\tabcolsep}{1pt} 

\newcommand{\resA}[2]{#1\textcolor{gray}{\tiny (#2)}}

\begin{tabularx}{\linewidth}{@{} l | *{12}{Y} @{}} 
\toprule
& \multicolumn{12}{c}{\textbf{Open-Source Models}} \\
\cmidrule(lr){2-13}
& \multicolumn{4}{c|}{\makecell{Qwen2.5-VL-7B-Instruct-AWQ \\ \scriptsize \citep{Qwen2.5-VL}}} & \multicolumn{4}{c|}{\makecell{Qwen2.5-VL-32B-Instruct-AWQ \\ \scriptsize \citep{Qwen2.5-VL}}} & \multicolumn{4}{c}{\makecell{InternVL2.5-8B \\ \scriptsize \citep{chen2024internvl}}} \\
Method & Base & CS & CI & VA & Base & CS & CI & VA & Base & CS & CI & VA \\
\midrule
Base Model    & \resA{32.7}{1.5} & \resA{30.7}{0.6} & \resA{32.0}{1.2} & \resA{27.3}{2.4} & \resA{49.3}{1.1} & \resA{43.3}{1.9} & \resA{46.7}{1.3} & \resA{43.3}{0.9} & \resA{29.3}{1.1} & \resA{22.7}{1.1} & \resA{24.0}{1.1} & \resA{21.3}{1.5} \\
Ablation1     & \resA{30.7}{1.1} & \resA{28.0}{0.7} & \resA{34.7}{1.8} & \resA{31.3}{1.5} & \resA{50.7}{0.6} & \resA{45.3}{1.5} & \resA{49.3}{1.7} & \resA{45.3}{0.7} & \resA{31.3}{0.7} & \resA{25.3}{1.5} & \resA{26.0}{1.5} & \resA{23.3}{0.9} \\
Ablation2     & \resA{34.0}{2.2} & \resA{31.3}{0.7} & \resA{32.7}{0.6} & \resA{30.0}{0.0} & \resA{51.3}{1.2} & \resA{46.7}{0.9} & \resA{48.0}{0.7} & \resA{45.3}{0.7} & \resA{32.7}{0.6} & \resA{27.3}{1.1} & \resA{28.7}{0.7} & \resA{22.7}{1.1} \\
Ablation3     & \resA{37.3}{0.6} & \resA{31.3}{1.2} & \resA{34.7}{0.7} & \resA{32.0}{1.5} & \resA{52.0}{0.7} & \resA{46.7}{1.6} & \resA{48.7}{1.8} & \resA{46.0}{0.6} & \resA{37.3}{0.6} & \resA{28.7}{1.8} & \resA{27.3}{1.5} & \resA{23.3}{1.3} \\
\midrule
\textbf{Ours} & \textbf{\resA{44.0}{1.1}} & \textbf{\resA{36.0}{0.6}} & \textbf{\resA{37.3}{0.6}} & \textbf{\resA{35.3}{0.7}} & \textbf{\resA{56.7}{1.3}} & \textbf{\resA{52.7}{1.1}} & \textbf{\resA{53.3}{0.9}} & \textbf{\resA{49.3}{1.1}} & \textbf{\resA{39.3}{0.7}} & \textbf{\resA{32.7}{0.6}} & \textbf{\resA{33.3}{1.1}} & \textbf{\resA{25.3}{0.7}} \\

\bottomrule
\end{tabularx}

\vspace{1.2em}

\begin{tabularx}{\linewidth}{@{} l | *{12}{Y} @{}}
\toprule
& \multicolumn{12}{c}{\textbf{Closed-Source Models}} \\
\cmidrule(lr){2-13}
& \multicolumn{4}{c|}{\makecell{gpt-4o-mini \\ \scriptsize \citep{openai2024gpt4o}}} & \multicolumn{4}{c|}{\makecell{gemini-2.5-flash \\ \scriptsize \citep{comanici2025gemini25pushingfrontier}}} & \multicolumn{4}{c}{\makecell{doubao-seed-1.6-flash \\ \scriptsize \citep{guo2025seed15vltechnicalreport}}} \\
Method & Base & CS & CI & VA & Base & CS & CI & VA & Base & CS & CI & VA \\
\midrule
Base Model    & \resA{43.3}{1.6} & \resA{37.3}{2.4} & \resA{38.7}{1.5} & \resA{37.3}{1.7} & \resA{68.0}{1.1} & \resA{62.0}{0.7} & \resA{60.0}{0.9} & \resA{55.3}{1.2} & \resA{59.3}{1.1} & \resA{40.0}{1.3} & \resA{38.0}{2.2} & \resA{32.7}{1.5} \\
Ablation1     & \resA{44.0}{1.5} & \resA{38.7}{2.0} & \resA{40.7}{1.1} & \resA{39.3}{1.1} & \resA{68.7}{2.0} & \resA{62.7}{0.6} & \resA{61.3}{0.7} & \resA{56.0}{1.1} & \resA{60.7}{0.6} & \resA{41.3}{0.7} & \resA{39.3}{1.5} & \resA{36.0}{1.5} \\
Ablation2     & \resA{44.0}{1.5} & \resA{44.0}{1.1} & \resA{44.7}{0.7} & \resA{42.7}{0.6} & \resA{70.7}{0.6} & \resA{64.0}{0.6} & \resA{63.3}{1.9} & \resA{58.7}{0.7} & \resA{62.7}{1.1} & \resA{44.0}{1.1} & \resA{41.3}{1.8} & \resA{40.0}{1.6} \\
Ablation3     & \resA{46.0}{1.1} & \resA{47.3}{0.6} & \resA{42.7}{0.6} & \resA{44.0}{1.1} & \resA{72.7}{0.6} & \resA{63.3}{0.9} & \resA{63.3}{2.1} & \resA{57.3}{1.7} & \resA{63.3}{1.3} & \resA{46.0}{1.1} & \resA{40.7}{1.8} & \resA{40.7}{1.5} \\
\midrule
\textbf{Ours} & \textbf{\resA{50.0}{0.9}} & \textbf{\resA{53.3}{0.9}} & \textbf{\resA{49.3}{1.1}} & \textbf{\resA{49.3}{0.6}} & \textbf{\resA{74.7}{1.2}} & \textbf{\resA{65.3}{0.7}} & \textbf{\resA{66.0}{0.6}} & \textbf{\resA{62.0}{1.2}} & \textbf{\resA{66.0}{1.1}} & \textbf{\resA{48.0}{0.7}} & \textbf{\resA{48.7}{1.2}} & \textbf{\resA{44.7}{0.7}} \\
\bottomrule
\end{tabularx}
\end{table*}

\textbf{Scene Graph Stress Testing:} Given that the scene graph serves as the key part of our algorithm, we conducted stress tests specifically within the context of room navigation. Since our scene graph is domain-specific, large-scale node loss is rare in real-world inference; thus, we simulated such failures by manually dropping a subset of nodes to evaluate robustness and to examine whether the lost nodes can be recovered through fine-stage exploration.

As shown in Table \ref{tab:stress_test}, our method demonstrates consistently high recovery rates and minimal hallucination rates across varying levels of node sparsity. Specifically, the recovery rate remains above 96\% across all drop ratios (10\% to 50\%), while the hallucination rate and miss rate are both maintained below 2\%. These quantitative results indicate that even in the event of partial scene graph collapse, our two-stage verification mechanism ensures algorithmic stability. For detailed qualitative analysis and visualization of the recovery process, please refer to the Appendix.

\begin{table}[!htbp]
    \centering
    \small
    \caption{Performance of SceneDiver under varying Drop Ratios (DR) across 500 scenes. We report the Recovery Rate (RR), VLM Hallucination Rate (HR), and Miss Rate (MR) to evaluate the system's robustness.}
    \label{tab:stress_test}
    \setlength{\tabcolsep}{10pt} 
    \begin{tabular}{c c c c}
        \toprule
        \textbf{DR} & \textbf{RR} ($\uparrow$) & \textbf{HR} ($\downarrow$) & \textbf{MR} ($\downarrow$) \\
        \midrule
        10\% & 96.78 & 1.52 & 1.70 \\
        20\% & 98.01 & 1.23 & 0.76 \\
        30\% & 97.16 & 1.89 & 0.95 \\
        50\% & 97.69 & 1.02 & 1.29 \\
        \bottomrule
    \end{tabular}
\end{table}

\section{Conclusion}
In this work, we address a critical perception bottleneck in embodied vision-language decision making, manifested in object hallucination and perceptual errors. We propose SceneDiver, a coarse-to-fine focus plan generation method for VLMs. We break the perception bottleneck, by enabling VLMs to autonomously generate the focus plan, which can help VLMs to focus only on task-relevant objects in each decision-making step. To enable reactive control, we also design a lightweight adapter for distilling the deliberate focus ability into VLAs. Extensive experiments demonstrate that SceneDiver significantly reduces visual hallucinations for both VLMs and
VLAs. 

{\bf Limitations and future work}. While SceneDiver demonstrates strong performance across multiple benchmarks, several directions remain for future exploration. Although our stress tests show robustness to scene graph noise, the generation module could benefit from further optimization to handle extremely dynamic scenes with rapid object motion. Additionally, while our method advances in long-horizon tasks, exploring its integration with hierarchical planning frameworks could potentially yield additional benefits for more complex multi-stage manipulation scenarios.

\section*{Acknowledgement}
This work was supported by National Natural Science Foundation of China (U23A20311, 62206245). We would like to thank the anonymous reviewers for their constructive suggestions.

\section*{Impact Statement}

This paper presents work whose goal is to advance the field of Machine
Learning. There are many potential societal consequences of our work, none of
which we feel must be specifically highlighted here.


\bibliography{conference}

@inproceedings{anderson2016spice,
  title={Spice: Semantic propositional image caption evaluation},
  author={Anderson, Peter and Fernando, Basura and Johnson, Mark and Gould, Stephen},
  booktitle={European conference on computer vision},
  pages={382--398},
  year={2016},
  organization={Springer}
}

@article{samuelson2025terra,
  title={Terra: Hierarchical Terrain-Aware 3D Scene Graph for Task-Agnostic Outdoor Mapping},
  author={Samuelson, Chad R and Austin, Abigail and Knoop, Seth and Romrell, Blake and Slade, Gabriel R and McLain, Timothy W and Mangelson, Joshua G},
  journal={arXiv preprint arXiv:2509.19579},
  year={2025}
}

@inproceedings{nguyen2025generating,
  title={Generating Actionable Robot Knowledge Bases by Combining 3D Scene Graphs with Robot Ontologies},
  author={Nguyen, Giang and Pomarlan, Mihai and Jongebloed, Sascha and Leusmann, Nils and Vu, Minh Nhat and Beetz, Michael},
  booktitle={2025 IEEE/RSJ International Conference on Intelligent Robots and Systems (IROS)},
  pages={21527--21534},
  year={2025},
  organization={IEEE}
}

@inproceedings{wang2019role,
  title={On the role of scene graphs in image captioning},
  author={Wang, Dalin and Beck, Daniel and Cohn, Trevor},
  booktitle={Proceedings of the beyond vision and language: integrating real-world knowledge (LANTERN)},
  pages={29--34},
  year={2019}
}

@inproceedings{yang2019auto,
  title={Auto-encoding scene graphs for image captioning},
  author={Yang, Xu and Tang, Kaihua and Zhang, Hanwang and Cai, Jianfei},
  booktitle={Proceedings of the IEEE/CVF conference on computer vision and pattern recognition},
  pages={10685--10694},
  year={2019}
}

@inproceedings{hudson2019gqa,
  title={Gqa: A new dataset for real-world visual reasoning and compositional question answering},
  author={Hudson, Drew A and Manning, Christopher D},
  booktitle={Proceedings of the IEEE/CVF conference on computer vision and pattern recognition},
  pages={6700--6709},
  year={2019}
}

@article{krishna2017visual,
  title={Visual genome: Connecting language and vision using crowdsourced dense image annotations},
  author={Krishna, Ranjay and Zhu, Yuke and Groth, Oliver and Johnson, Justin and Hata, Kenji and Kravitz, Joshua and Chen, Stephanie and Kalantidis, Yannis and Li, Li-Jia and Shamma, David A and others},
  journal={International journal of computer vision},
  volume={123},
  number={1},
  pages={32--73},
  year={2017},
  publisher={Springer}
}

@inproceedings{johnson2015image,
  title={Image retrieval using scene graphs},
  author={Johnson, Justin and Krishna, Ranjay and Stark, Michael and Li, Li-Jia and Shamma, David and Bernstein, Michael and Fei-Fei, Li},
  booktitle={Proceedings of the IEEE conference on computer vision and pattern recognition},
  pages={3668--3678},
  year={2015}
}

@misc{lin2023sphinxjointmixingweights,
      title={SPHINX: The Joint Mixing of Weights, Tasks, and Visual Embeddings for Multi-modal Large Language Models}, 
      author={Ziyi Lin and Chris Liu and Renrui Zhang and Peng Gao and Longtian Qiu and Han Xiao and Han Qiu and Chen Lin and Wenqi Shao and Keqin Chen and Jiaming Han and Siyuan Huang and Yichi Zhang and Xuming He and Hongsheng Li and Yu Qiao},
      year={2023},
      eprint={2311.07575},
      archivePrefix={arXiv},
      primaryClass={cs.CV},
      url={https://arxiv.org/abs/2311.07575}, 
}

@misc{li2024llavaonevisioneasyvisualtask,
      title={LLaVA-OneVision: Easy Visual Task Transfer}, 
      author={Bo Li and Yuanhan Zhang and Dong Guo and Renrui Zhang and Feng Li and Hao Zhang and Kaichen Zhang and Peiyuan Zhang and Yanwei Li and Ziwei Liu and Chunyuan Li},
      year={2024},
      eprint={2408.03326},
      archivePrefix={arXiv},
      primaryClass={cs.CV},
      url={https://arxiv.org/abs/2408.03326}, 
}

@article{kuhn1955hungarian,
  title={The Hungarian method for the assignment problem},
  author={Kuhn, Harold W},
  journal={Naval research logistics quarterly},
  volume={2},
  number={1-2},
  pages={83--97},
  year={1955},
  publisher={Wiley Online Library}
}

@article{sapkota2025vision,
  title={Vision-language-action models: Concepts, progress, applications and challenges},
  author={Sapkota, Ranjan and Cao, Yang and Roumeliotis, Konstantinos I and Karkee, Manoj},
  journal={arXiv preprint arXiv:2505.04769},
  year={2025}
}

@misc{brohan2023rt2visionlanguageactionmodelstransfer,
      title={RT-2: Vision-Language-Action Models Transfer Web Knowledge to Robotic Control}, 
      author={Anthony Brohan and Noah Brown and Justice Carbajal and Yevgen Chebotar and Xi Chen and Krzysztof Choromanski and Tianli Ding and Danny Driess and Avinava Dubey and Chelsea Finn and Pete Florence and Chuyuan Fu and Montse Gonzalez Arenas and Keerthana Gopalakrishnan and Kehang Han and Karol Hausman and Alexander Herzog and Jasmine Hsu and Brian Ichter and Alex Irpan and Nikhil Joshi and Ryan Julian and Dmitry Kalashnikov and Yuheng Kuang and Isabel Leal and Lisa Lee and Tsang-Wei Edward Lee and Sergey Levine and Yao Lu and Henryk Michalewski and Igor Mordatch and Karl Pertsch and Kanishka Rao and Krista Reymann and Michael Ryoo and Grecia Salazar and Pannag Sanketi and Pierre Sermanet and Jaspiar Singh and Anikait Singh and Radu Soricut and Huong Tran and Vincent Vanhoucke and Quan Vuong and Ayzaan Wahid and Stefan Welker and Paul Wohlhart and Jialin Wu and Fei Xia and Ted Xiao and Peng Xu and Sichun Xu and Tianhe Yu and Brianna Zitkovich},
      year={2023},
      eprint={2307.15818},
      archivePrefix={arXiv},
      primaryClass={cs.RO},
      url={https://arxiv.org/abs/2307.15818}, 
}

@misc{driess2023palmeembodiedmultimodallanguage,
      title={PaLM-E: An Embodied Multimodal Language Model}, 
      author={Danny Driess and Fei Xia and Mehdi S. M. Sajjadi and Corey Lynch and Aakanksha Chowdhery and Brian Ichter and Ayzaan Wahid and Jonathan Tompson and Quan Vuong and Tianhe Yu and Wenlong Huang and Yevgen Chebotar and Pierre Sermanet and Daniel Duckworth and Sergey Levine and Vincent Vanhoucke and Karol Hausman and Marc Toussaint and Klaus Greff and Andy Zeng and Igor Mordatch and Pete Florence},
      year={2023},
      eprint={2303.03378},
      archivePrefix={arXiv},
      primaryClass={cs.LG},
      url={https://arxiv.org/abs/2303.03378}, 
}

@misc{ahn2022icanisay,
      title={Do As I Can, Not As I Say: Grounding Language in Robotic Affordances}, 
      author={Michael Ahn and Anthony Brohan and Noah Brown and Yevgen Chebotar and Omar Cortes and Byron David and Chelsea Finn and Chuyuan Fu and Keerthana Gopalakrishnan and Karol Hausman and Alex Herzog and Daniel Ho and Jasmine Hsu and Julian Ibarz and Brian Ichter and Alex Irpan and Eric Jang and Rosario Jauregui Ruano and Kyle Jeffrey and Sally Jesmonth and Nikhil J Joshi and Ryan Julian and Dmitry Kalashnikov and Yuheng Kuang and Kuang-Huei Lee and Sergey Levine and Yao Lu and Linda Luu and Carolina Parada and Peter Pastor and Jornell Quiambao and Kanishka Rao and Jarek Rettinghouse and Diego Reyes and Pierre Sermanet and Nicolas Sievers and Clayton Tan and Alexander Toshev and Vincent Vanhoucke and Fei Xia and Ted Xiao and Peng Xu and Sichun Xu and Mengyuan Yan and Andy Zeng},
      year={2022},
      eprint={2204.01691},
      archivePrefix={arXiv},
      primaryClass={cs.RO},
      url={https://arxiv.org/abs/2204.01691}, 
}

@misc{leng2023mitigatingobjecthallucinationslarge,
      title={Mitigating Object Hallucinations in Large Vision-Language Models through Visual Contrastive Decoding}, 
      author={Sicong Leng and Hang Zhang and Guanzheng Chen and Xin Li and Shijian Lu and Chunyan Miao and Lidong Bing},
      year={2023},
      eprint={2311.16922},
      archivePrefix={arXiv},
      primaryClass={cs.CV},
      url={https://arxiv.org/abs/2311.16922}, 
}

@misc{li2024monkeyimageresolutiontext,
      title={Monkey: Image Resolution and Text Label Are Important Things for Large Multi-modal Models}, 
      author={Zhang Li and Biao Yang and Qiang Liu and Zhiyin Ma and Shuo Zhang and Jingxu Yang and Yabo Sun and Yuliang Liu and Xiang Bai},
      year={2024},
      eprint={2311.06607},
      archivePrefix={arXiv},
      primaryClass={cs.CV},
      url={https://arxiv.org/abs/2311.06607}, 
}

@misc{yang2023setofmarkpromptingunleashesextraordinary,
      title={Set-of-Mark Prompting Unleashes Extraordinary Visual Grounding in GPT-4V}, 
      author={Jianwei Yang and Hao Zhang and Feng Li and Xueyan Zou and Chunyuan Li and Jianfeng Gao},
      year={2023},
      eprint={2310.11441},
      archivePrefix={arXiv},
      primaryClass={cs.CV},
      url={https://arxiv.org/abs/2310.11441}, 
}

@misc{black2026pi0visionlanguageactionflowmodel,
      title={$\pi_0$: A Vision-Language-Action Flow Model for General Robot Control}, 
      author={Kevin Black and Noah Brown and Danny Driess and Adnan Esmail and Michael Equi and Chelsea Finn and Niccolo Fusai and Lachy Groom and Karol Hausman and Brian Ichter and Szymon Jakubczak and Tim Jones and Liyiming Ke and Sergey Levine and Adrian Li-Bell and Mohith Mothukuri and Suraj Nair and Karl Pertsch and Lucy Xiaoyang Shi and James Tanner and Quan Vuong and Anna Walling and Haohuan Wang and Ury Zhilinsky},
      year={2026},
      eprint={2410.24164},
      archivePrefix={arXiv},
      primaryClass={cs.LG},
      url={https://arxiv.org/abs/2410.24164}, 
}

@misc{fei2025liberoplusindepthrobustnessanalysis,
      title={LIBERO-Plus: In-depth Robustness Analysis of Vision-Language-Action Models}, 
      author={Senyu Fei and Siyin Wang and Junhao Shi and Zihao Dai and Jikun Cai and Pengfang Qian and Li Ji and Xinzhe He and Shiduo Zhang and Zhaoye Fei and Jinlan Fu and Jingjing Gong and Xipeng Qiu},
      year={2025},
      eprint={2510.13626},
      archivePrefix={arXiv},
      primaryClass={cs.RO},
      url={https://arxiv.org/abs/2510.13626}, 
}

@misc{locatello2020objectcentriclearningslotattention,
      title={Object-Centric Learning with Slot Attention}, 
      author={Francesco Locatello and Dirk Weissenborn and Thomas Unterthiner and Aravindh Mahendran and Georg Heigold and Jakob Uszkoreit and Alexey Dosovitskiy and Thomas Kipf},
      year={2020},
      eprint={2006.15055},
      archivePrefix={arXiv},
      primaryClass={cs.LG},
      url={https://arxiv.org/abs/2006.15055}, 
}

@misc{kim2025finetuningvisionlanguageactionmodelsoptimizing,
      title={Fine-Tuning Vision-Language-Action Models: Optimizing Speed and Success}, 
      author={Moo Jin Kim and Chelsea Finn and Percy Liang},
      year={2025},
      eprint={2502.19645},
      archivePrefix={arXiv},
      primaryClass={cs.RO},
      url={https://arxiv.org/abs/2502.19645}, 
}

@misc{chen2024expandingscenegraphboundaries,
      title={Expanding Scene Graph Boundaries: Fully Open-vocabulary Scene Graph Generation via Visual-Concept Alignment and Retention}, 
      author={Zuyao Chen and Jinlin Wu and Zhen Lei and Zhaoxiang Zhang and Changwen Chen},
      year={2024},
      eprint={2311.10988},
      archivePrefix={arXiv},
      primaryClass={cs.CV},
      url={https://arxiv.org/abs/2311.10988}, 
}

@misc{yang2025embodiedbenchcomprehensivebenchmarkingmultimodal,
      title={EmbodiedBench: Comprehensive Benchmarking Multi-modal Large Language Models for Vision-Driven Embodied Agents}, 
      author={Rui Yang and Hanyang Chen and Junyu Zhang and Mark Zhao and Cheng Qian and Kangrui Wang and Qineng Wang and Teja Venkat Koripella and Marziyeh Movahedi and Manling Li and Heng Ji and Huan Zhang and Tong Zhang},
      year={2025},
      eprint={2502.09560},
      archivePrefix={arXiv},
      primaryClass={cs.AI},
      url={https://arxiv.org/abs/2502.09560}, 
}

@misc{reflectVLM,
      title={Reflective Planning: Vision-Language Models for Multi-Stage Long-Horizon Robotic Manipulation}, 
      author={Yunhai Feng and Jiaming Han and Zhuoran Yang and Xiangyu Yue and Sergey Levine and Jianlan Luo},
      year={2025},
      eprint={2502.16707},
      archivePrefix={arXiv},
      primaryClass={cs.RO},
      url={https://arxiv.org/abs/2502.16707}, 
}

@article{ma2024survey,
  title={A survey on vision-language-action models for embodied ai},
  author={Ma, Yueen and Song, Zixing and Zhuang, Yuzheng and Hao, Jianye and King, Irwin},
  journal={arXiv preprint arXiv:2405.14093},
  year={2024}
}

@inproceedings{brohan2023can,
  title={Do as i can, not as i say: Grounding language in robotic affordances},
  author={Brohan, Anthony and Chebotar, Yevgen and Finn, Chelsea and Hausman, Karol and Herzog, Alexander and Ho, Daniel and Ibarz, Julian and Irpan, Alex and Jang, Eric and Julian, Ryan and others},
  booktitle={Conference on robot learning},
  pages={287--318},
  year={2023},
  organization={PMLR}
}

@inproceedings{huang2022language,
  title={Language models as zero-shot planners: Extracting actionable knowledge for embodied agents},
  author={Huang, Wenlong and Abbeel, Pieter and Pathak, Deepak and Mordatch, Igor},
  booktitle={International conference on machine learning},
  pages={9118--9147},
  year={2022},
  organization={PMLR}
}

@inproceedings{jang2022bc,
  title={Bc-z: Zero-shot task generalization with robotic imitation learning},
  author={Jang, Eric and Irpan, Alex and Khansari, Mohi and Kappler, Daniel and Ebert, Frederik and Lynch, Corey and Levine, Sergey and Finn, Chelsea},
  booktitle={Conference on Robot Learning},
  pages={991--1002},
  year={2022},
  organization={PMLR}
}

@article{shang2024theia,
  title={Theia: Distilling diverse vision foundation models for robot learning},
  author={Shang, Jinghuan and Schmeckpeper, Karl and May, Brandon B and Minniti, Maria Vittoria and Kelestemur, Tarik and Watkins, David and Herlant, Laura},
  journal={arXiv preprint arXiv:2407.20179},
  year={2024}
}

@inproceedings{assran2023self,
  title={Self-supervised learning from images with a joint-embedding predictive architecture},
  author={Assran, Mahmoud and Duval, Quentin and Misra, Ishan and Bojanowski, Piotr and Vincent, Pascal and Rabbat, Michael and LeCun, Yann and Ballas, Nicolas},
  booktitle={Proceedings of the IEEE/CVF Conference on Computer Vision and Pattern Recognition},
  pages={15619--15629},
  year={2023}
}

@article{karamcheti2023language,
  title={Language-driven representation learning for robotics},
  author={Karamcheti, Siddharth and Nair, Suraj and Chen, Annie S and Kollar, Thomas and Finn, Chelsea and Sadigh, Dorsa and Liang, Percy},
  journal={arXiv preprint arXiv:2302.12766},
  year={2023}
}

@misc{rohrbach2019objecthallucinationimagecaptioning,
      title={Object Hallucination in Image Captioning}, 
      author={Anna Rohrbach and Lisa Anne Hendricks and Kaylee Burns and Trevor Darrell and Kate Saenko},
      year={2019},
      eprint={1809.02156},
      archivePrefix={arXiv},
      primaryClass={cs.CL},
      url={https://arxiv.org/abs/1809.02156}, 
}

@misc{dai2023plausiblefaithfulprobingobject,
      title={Plausible May Not Be Faithful: Probing Object Hallucination in Vision-Language Pre-training}, 
      author={Wenliang Dai and Zihan Liu and Ziwei Ji and Dan Su and Pascale Fung},
      year={2023},
      eprint={2210.07688},
      archivePrefix={arXiv},
      primaryClass={cs.CL},
      url={https://arxiv.org/abs/2210.07688}, 
}

@misc{zhai2024hallecontrolcontrollingobjecthallucination,
      title={HallE-Control: Controlling Object Hallucination in Large Multimodal Models}, 
      author={Bohan Zhai and Shijia Yang and Chenfeng Xu and Sheng Shen and Kurt Keutzer and Chunyuan Li and Manling Li},
      year={2024},
      eprint={2310.01779},
      archivePrefix={arXiv},
      primaryClass={cs.CV},
      url={https://arxiv.org/abs/2310.01779}, 
}

@misc{you2023ferretrefergroundgranularity,
      title={Ferret: Refer and Ground Anything Anywhere at Any Granularity}, 
      author={Haoxuan You and Haotian Zhang and Zhe Gan and Xianzhi Du and Bowen Zhang and Zirui Wang and Liangliang Cao and Shih-Fu Chang and Yinfei Yang},
      year={2023},
      eprint={2310.07704},
      archivePrefix={arXiv},
      primaryClass={cs.CV},
      url={https://arxiv.org/abs/2310.07704}, 
}

@misc{zhang2024groundhoggroundinglargelanguage,
      title={GROUNDHOG: Grounding Large Language Models to Holistic Segmentation}, 
      author={Yichi Zhang and Ziqiao Ma and Xiaofeng Gao and Suhaila Shakiah and Qiaozi Gao and Joyce Chai},
      year={2024},
      eprint={2402.16846},
      archivePrefix={arXiv},
      primaryClass={cs.CV},
      url={https://arxiv.org/abs/2402.16846}, 
}

@misc{sun2023aligninglargemultimodalmodels,
      title={Aligning Large Multimodal Models with Factually Augmented RLHF}, 
      author={Zhiqing Sun and Sheng Shen and Shengcao Cao and Haotian Liu and Chunyuan Li and Yikang Shen and Chuang Gan and Liang-Yan Gui and Yu-Xiong Wang and Yiming Yang and Kurt Keutzer and Trevor Darrell},
      year={2023},
      eprint={2309.14525},
      archivePrefix={arXiv},
      primaryClass={cs.CV},
      url={https://arxiv.org/abs/2309.14525}, 
}

@misc{gunjal2024detectingpreventinghallucinationslarge,
      title={Detecting and Preventing Hallucinations in Large Vision Language Models}, 
      author={Anisha Gunjal and Jihan Yin and Erhan Bas},
      year={2024},
      eprint={2308.06394},
      archivePrefix={arXiv},
      primaryClass={cs.CV},
      url={https://arxiv.org/abs/2308.06394}, 
}

@article{li2025survey,
  title={A survey of state of the art large vision language models: Alignment, benchmark, evaluations and challenges},
  author={Li, Zongxia and Wu, Xiyang and Du, Hongyang and Liu, Fuxiao and Nghiem, Huy and Shi, Guangyao},
  journal={arXiv preprint arXiv:2501.02189},
  year={2025}
}

@inproceedings{danish2025geobenchvlm,
  title     = {GEOBench-VLM: Benchmarking Vision-Language Models for Geospatial Tasks},
  author    = {Muhammad Sohail Danish and Muhammad Akhtar Munir and Syed Roshaan Ali Shah and Kartik Kuckreja and Fahad Shahbaz Khan and Paolo Fraccaro and Alexandre Lacoste and Salman Khan},
  booktitle = {Proceedings of the IEEE/CVF International Conference on Computer Vision (ICCV)},
  year      = {2025},
  url       = {https://arxiv.org/abs/2411.19325},
  archivePrefix = {arXiv},
  eprint    = {2411.19325},
  primaryClass = {cs.CV}
}

@article{zong2024vlicl,
  title={VL-ICL Bench: The Devil in the Details of Multimodal In-Context Learning},
  author={Zong, Yongshuo and Bohdal, Ondrej and Hospedales, Timothy},
  journal={arXiv preprint arXiv:2403.13164},
  year={2024}
}

@inproceedings{chen2024spatialvlm,
  title={Spatialvlm: Endowing vision-language models with spatial reasoning capabilities},
  author={Chen, Boyuan and Xu, Zhuo and Kirmani, Sean and Ichter, Brain and Sadigh, Dorsa and Guibas, Leonidas and Xia, Fei},
  booktitle={Proceedings of the IEEE/CVF Conference on Computer Vision and Pattern Recognition},
  pages={14455--14465},
  year={2024}
}

@misc{liu2023llava,
      title={Visual Instruction Tuning}, 
      author={Liu, Haotian and Li, Chunyuan and Wu, Qingyang and Lee, Yong Jae},
      publisher={NeurIPS},
      year={2023},
}

@inproceedings{Multi-object-hallucination,
    author = {Chen, Xuweiyi and Ma, Ziqiao and Zhang, Xuejun and Xu, Sihan and Qian, Shengyi and Yang, Jianing and Fouhey, David F. and Chai, Joyce},
    title = {Multi-object hallucination in vision language models},
    year = {2025},
    isbn = {9798331314385},
    publisher = {Curran Associates Inc.},
    address = {Red Hook, NY, USA},
    articleno = {1409},
    numpages = {26},
    location = {Vancouver, BC, Canada},
    series = {NIPS '24}
}

@article{zhu2025mitigating,
  title={Mitigating object hallucinations in large vision-language models via attention calibration},
  author={Zhu, Younan and Tao, Linwei and Dong, Minjing and Xu, Chang},
  journal={arXiv preprint arXiv:2502.01969},
  year={2025}
}

@inproceedings{huang2024opera,
  title={Opera: Alleviating hallucination in multi-modal large language models via over-trust penalty and retrospection-allocation},
  author={Huang, Qidong and Dong, Xiaoyi and Zhang, Pan and Wang, Bin and He, Conghui and Wang, Jiaqi and Lin, Dahua and Zhang, Weiming and Yu, Nenghai},
  booktitle={Proceedings of the IEEE/CVF Conference on Computer Vision and Pattern Recognition},
  pages={13418--13427},
  year={2024}
}

@inproceedings{todorov2012mujoco,
  title={MuJoCo: A physics engine for model-based control},
  author={Todorov, Emanuel and Erez, Tom and Tassa, Yuval},
  booktitle={2012 IEEE/RSJ International Conference on Intelligent Robots and Systems},
  pages={5026--5033},
  year={2012},
  organization={IEEE},
  doi={10.1109/IROS.2012.6386109}
}

@article{Qwen2.5-VL,
  title={Qwen2.5-VL Technical Report},
  author={Bai, Shuai and Chen, Keqin and Liu, Xuejing and Wang, Jialin and Ge, Wenbin and Song, Sibo and Dang, Kai and Wang, Peng and Wang, Shijie and Tang, Jun and Zhong, Humen and Zhu, Yuanzhi and Yang, Mingkun and Li, Zhaohai and Wan, Jianqiang and Wang, Pengfei and Ding, Wei and Fu, Zheren and Xu, Yiheng and Ye, Jiabo and Zhang, Xi and Xie, Tianbao and Cheng, Zesen and Zhang, Hang and Yang, Zhibo and Xu, Haiyang and Lin, Junyang},
  journal={arXiv preprint arXiv:2502.13923},
  year={2025}
}

@inproceedings{chen2024internvl,
  title={Internvl: Scaling up vision foundation models and aligning for generic visual-linguistic tasks},
  author={Chen, Zhe and Wu, Jiannan and Wang, Wenhai and Su, Weijie and Chen, Guo and Xing, Sen and Zhong, Muyan and Zhang, Qinglong and Zhu, Xizhou and Lu, Lewei and others},
  booktitle={Proceedings of the IEEE/CVF Conference on Computer Vision and Pattern Recognition},
  pages={24185--24198},
  year={2024}
}

@misc{guo2025seed15vltechnicalreport,
      title={Seed1.5-VL Technical Report}, 
      author={Dong Guo and et al.},
      year={2025},
      eprint={2505.07062},
      archivePrefix={arXiv},
      primaryClass={cs.CV},
      url={https://arxiv.org/abs/2505.07062}, 
}

@misc{comanici2025gemini25pushingfrontier,
      title={Gemini 2.5: Pushing the Frontier with Advanced Reasoning, Multimodality, Long Context, and Next Generation Agentic Capabilities}, 
      author={Gheorghe Comanici and et al.},
      year={2025},
      eprint={2507.06261},
      archivePrefix={arXiv},
      primaryClass={cs.CL},
      url={https://arxiv.org/abs/2507.06261}, 
}

@misc{openai2024gpt4o,
  author = {OpenAI},
  title = {Hello GPT-4o},
  year = {2024},
  url = {https://openai.com/index/hello-gpt-4o/},
  urldate = {2024-05-13},
  note = {Accessed: 2024-05-13}
}
\bibliographystyle{icml2026}

\newpage

\appendix
\onecolumn
\section{Appendix}
\subsection{Extended Ablation Studies}
To evaluate the role of the adapter, we compare it with an end-to-end model that predicts the focus directly from the input image. The experimental setup and evaluation metrics are consistent with those in the main text. The results are as follows: 

\begin{table*}[!htbp]
    \centering
    \caption{Performance comparison on LIBERO-plus benchmark suites. In each cell, the bold black values represent our method, while the smaller grey values below represent the end-to-end baseline. We report the mean success rate and standard error over 3 seeds.}
    \label{tab:libero_ablation}
    \renewcommand{\arraystretch}{1.4} 
    \setlength{\tabcolsep}{3pt}       
    \begin{tabular}{lccccc}
        \toprule
        \textbf{Task Suite} & \textbf{Objects Layout} & \textbf{Camera Viewpoints} & \textbf{Robot Initial States} & \textbf{Background Textures} & \textbf{Light Conditions} \\
        \midrule
        
        spatial & \res{93.21}{0.27}{94.43}{0.41}{1.22} & \res{51.54}{0.71}{54.35}{0.99}{2.81} & \res{23.45}{0.15}{29.49}{0.30}{6.04} & \res{86.41}{0.20}{87.41}{0.40}{1.00} & \res{96.31}{0.53}{97.37}{0.88}{1.06} \\
        \midrule
        
        object & \res{76.41}{0.13}{76.82}{0.27}{0.41} & \res{63.36}{0.68}{64.56}{0.86}{1.20} & \res{14.90}{0.00}{15.72}{0.00}{0.82} & \res{90.18}{0.66}{90.61}{1.53}{0.43} & \res{99.33}{0.00}{99.50}{0.17}{0.17} \\
        \midrule
        
        goal & \res{50.51}{0.38}{52.53}{0.38}{2.02} & \res{51.59}{0.43}{52.74}{0.72}{1.15} & \res{14.14}{0.00}{18.85}{0.00}{4.71} & \res{88.46}{0.61}{91.09}{1.21}{2.63} & \res{96.59}{0.38}{97.35}{0.76}{0.76} \\
        \midrule
        
        long & \res{68.49}{0.94}{74.52}{0.56}{6.03} & \res{41.22}{0.40}{42.97}{0.54}{1.75} & \res{34.17}{0.40}{37.73}{0.00}{3.56} & \res{86.97}{0.54}{90.00}{0.00}{3.03} & \res{89.96}{0.57}{92.42}{0.00}{2.46} \\
        
        \bottomrule
    \end{tabular}
\end{table*}

Compared with the direct end-to-end baseline, our method performs better in all 20 LIBERO-plus settings, with larger gains under harder distribution shifts, including robot initial states, object layouts, and camera viewpoints.

\subsection{Qualitative Analysis of Fine-grained Verification}
\label{sec:qualitative_verification}
\begin{figure*}[!htbp]
    \centering  
    \includegraphics[width=0.9\linewidth]{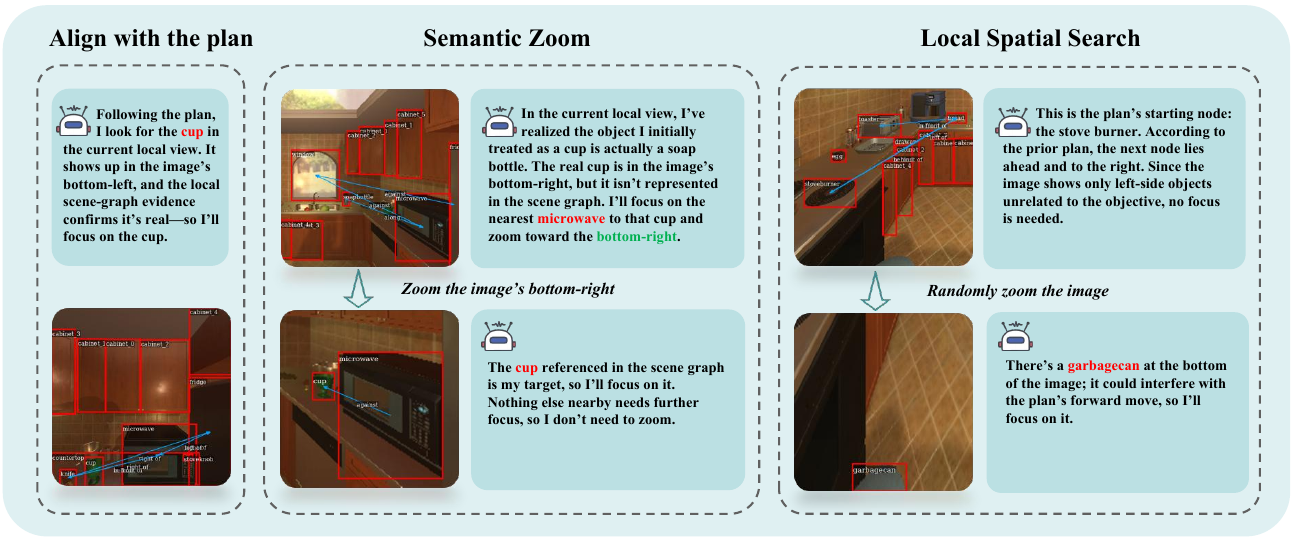} 
    \caption{Fine-grained Verification and Exploration results of three different agentic behaviors} 
    \label{fig:verification}   
\end{figure*}

In the fine-grained exploration stage, we perform a zoom-in operation on each candidate node to construct a local scene graph, recovering details that may be invisible in the global view. Based on the local observations, the VLM adaptively executes one of three exploration primitives, as illustrated in Figure~\ref{fig:verification}.

\noindent\textbf{Behavior 1.} When the target object is successfully detected within the local scene graph, the VLM performs multi-scale consistency verification to confirm that the object strictly meets the task requirements. This primitive encompasses two scenarios: (a) the target was identified in the coarse stage and is now re-confirmed locally, or (b) the coarse plan relied on a spatial proxy due to initial detection failures, and the zoom-in operation successfully reveals the originally missed target. Upon successful verification, the object is added to the final candidate set.

\noindent\textbf{Behavior 2.} When the target remains undetected after zoom-in—either because the local scene graph fails to capture it, or the detection results contradict the VLM's analysis—but a valid spatial sub-scene exists, the VLM leverages nearby objects in the scene graph as proxies. Using their coordinates, the model performs a further zoom into specific regions for deeper inspection.

\noindent\textbf{Behavior 3.} When the target remains undetected and no clear sub-scene is available from the scene graph, the VLM resorts to the overall visual context. It executes a directional zoom based on spatial heuristics to explore the neighborhood and recover implicit targets that are unmapped by the scene graph.

\subsection{Perceptual Error Analysis}
We quantitatively evaluated perception error rate in the navigation environment, including: missed present objects (Miss Rate), predicting nonexistent objects (False Positive Rate), wrong attribute binding (Bind Error Rate), incomplete same-class instance detection (Multi-Instance Error Rate), and missed task target (Target Miss Rate). For each sample, we determined whether each of these five error types occurred based on the consistency between the model output and the ground-truth scene annotations, and we report the proportion of each error type in both the original environment and the modified environment used in the main text. The results are shown in the table below:

\begin{table}[htbp]
\centering
\caption{Results of perceptual error rates, 5 different types are reported in percentage (\%).}
\label{tab:env_results}
\begin{tabular}{lccccc}
\toprule
\textbf{Qwen2.5-VL-7B-Instruct-AWQ} & \textbf{MR} & \textbf{FPR} & \textbf{BER} & \textbf{MIER} & \textbf{TMR} \\ 
\midrule
original env & 46.6 & 2.3 & 40.1 & 75.4 & 41.7 \\
modified env & 47.8 & 3.3 & 42.6 & 75.6 & 60.0 \\ 
\midrule
\midrule
\textbf{gemini-2.5-flash} & \textbf{MR} & \textbf{FPR} & \textbf{BER} & \textbf{MIER} & \textbf{TMR(\%)} \\ 
\midrule
original env & 23.8 & 4.6 & 57.8 & 77.0 & 22.6 \\
modified env & 31.9 & 2.9 & 53.3 & 74.6 & 42.9 \\ 
\bottomrule
\end{tabular}
\end{table}

\subsection{Experimental Results on the Vanilla Benchmark}
In this section, we evaluate our method on vanilla EmbodiedBench and then explain why we also use Mujoco and modify the navigation task.

As shown in Table ~\ref{tab:vanilla_navigation}, our method still outperforms the base model on the vanilla navigation task. We choose modified navigation task because it introduces conditions that are more representative of real-world deployment, where perception is frequently challenged by occlusion, clutter, and partial observability.

\begin{table}[h]
\centering
\caption{Result on Vanilla EmbodiedBench navigation task. Means and standard errors over 5 seeds are reported in percentage (\%).}
\label{tab:vanilla_navigation}
\begin{tabular}{lcccc}
\toprule
\textbf{Qwen2.5-VL-7B-Instruct-AWQ} & \textbf{base} & \textbf{common} & \textbf{complex} & \textbf{visual} \\ 
\midrule
base & $\text{36.3}_{(\text{1.4})}$ & $\text{25.0}_{(\text{0.5})}$ & $\text{35.0}_{(\text{1.2})}$ & $\text{28.3}_{(\text{1.2})}$ \\
ours & $\text{40.7}_{(\text{1.6})}$ & $\text{29.7}_{(\text{0.6})}$ & $\text{38.0}_{(\text{0.8})}$ & $\text{32.3}_{(\text{1.2})}$ \\
\midrule
\midrule
\textbf{gemini-2.5-flash} & \textbf{base} & \textbf{common} & \textbf{complex} & \textbf{visual} \\ 
\midrule
base & $\text{68.3}_{(\text{0.5})}$ & $\text{65.0}_{(\text{0.5})}$ & $\text{58.3}_{(\text{0.5})}$ & $\text{56.7}_{(\text{0.5})}$ \\
ours & $\text{71.7}_{(\text{0.5})}$ & $\text{67.7}_{(\text{0.4})}$ & $\text{61.0}_{(\text{0.4})}$ & $\text{61.0}_{(\text{0.4})}$ \\
\bottomrule
\end{tabular}
\end{table}

We choose MuJoCo as the first experimental platform because it better matches the problem setting of this paper and offers stronger controllability. We focus on the perception bottleneck of VLMs under distracting conditions, and MuJoCo allows us to directly inject controllable distractions into the scene, making it better suited for validating the core mechanism of our method. In contrast, vanilla EmbodiedBench Manipulation does not fully align with the problem setting studied in this paper. As noted in EmbodiedBench~\citep{yang2025embodiedbenchcomprehensivebenchmarkingmultimodal}, EB-Manipulation uses pre-given detection boxes and visual markers to help the model localize key objects. Such explicit visual cues substantially reduce the difficulty of perception, thereby weakening the perception bottleneck that this paper aims to investigate. To better match the problem setting studied in this paper, we removed the detection boxes while retaining only the axis information required for action execution. The experimental setup and evaluation metrics are consistent with those in the main text. As shown in the table below, our method still demonstrates a consistent improvement over the base model:

\begin{table}[h]
\centering
\caption{Result on Vanilla EmbodiedBench manipulation task. Means and standard errors over 5 seeds are reported in percentage (\%).}
\label{tab:vanilla_manipulation}
\begin{tabular}{lccccc}
\toprule
\textbf{Qwen2.5-VL-7B-Instruct-AWQ} & \textbf{base} & \textbf{common} & \textbf{complex} & \textbf{spatial} & \textbf{visual} \\ 
\midrule
base & $\text{6.7}_{(\text{0.8})}$ & $\text{8.8}_{(\text{0.8})}$ & $\text{10.4}_{(\text{0.7})}$ & $\text{15.0}_{(\text{0.8})}$ & $\text{8.3}_{(\text{0.9})}$ \\
ours & $\text{12.5}_{(\text{0.7})}$ & $\text{10.4}_{(\text{0.7})}$ & $\text{12.9}_{(\text{0.8})}$ & $\text{16.3}_{(\text{0.8})}$ & $\text{16.7}_{(\text{0.9})}$ \\ 
\midrule
\midrule
\textbf{gemini-2.5-flash} & \textbf{base} & \textbf{common} & \textbf{complex} & \textbf{spatial} & \textbf{visual} \\ 
\midrule
base & $\text{20.8}_{(\text{0.7})}$ & $\text{14.6}_{(\text{0.7})}$ & $\text{22.9}_{(\text{0.7})}$ & $\text{24.2}_{(\text{0.5})}$ & $\text{23.3}_{(\text{0.7})}$ \\
ours & $\text{28.3}_{(\text{0.5})}$ & $\text{17.9}_{(\text{0.5})}$ & $\text{26.3}_{(\text{0.5})}$ & $\text{28.3}_{(\text{0.5})}$ & $\text{33.3}_{(\text{0.9})}$ \\
\bottomrule
\end{tabular}
\end{table}

\subsection{Inference Latency Breakdown of SceneDiver Components}
\label{sec:latency}

\begin{table}[h]
\centering
\caption{Detailed Latency Profiling and Tail Latency Analysis (ms)}
\label{tab:detailed_latency}
\begin{tabular}{lccccc}
\hline
\textbf{Module} & \textbf{Mean} & \textbf{Median} & \textbf{P95} & \textbf{P99} & \textbf{Percentage} \\ \hline
Action (Total) & 114.45 & 113.81 & 116.50 & 121.78 & 100.00\% \\
SlotAttention           & 2.18   & 2.16   & 2.30   & 2.33   & 1.91\%   \\
MaskNet             & 0.83   & 0.83   & 0.86   & 0.87   & 0.73\%   \\ \hline
\end{tabular}
\end{table}

To evaluate the computational impact of the SceneDiver adapter on the OpenVLA-OFT model, we measured inference latency on a single NVIDIA RTX 4090 GPU. We benchmarked the latency of the Slot Attention and MaskNet modules over a sequence of 10,000 action generations. As shown in the table, the overhead introduced by SceneDiver (approximately 3.01ms) accounts for only 2.64\% of the total action generation time (114.45ms). Crucially, the P99 latency of our modules remains within a very tight bound, ensuring the real-time reliability and determinism of the robotic system even in cluttered, multi-object environments.

\subsection{Implementation Details}
\subsubsection{SceneGraphGeneration}
We adopt the OvSGTR algorithm proposed by \citet{chen2024expandingscenegraphboundaries} as our primary scene-graph generator. Benefiting from its strong inherent generalization capabilities, the model achieves effective performance with a limited number of training samples. Our training dataset is synthesized from randomized scenes: for each scene, we extract 3D coordinates and semantic labels, then project these 3D points onto the 2D image plane using camera pose data. Finally, the model is trained using the official OvSGTR training protocols.

During the data generation process, each frame yields three synchronized outputs designed for scene-graph supervision:
\begin{itemize}
    \item \textbf{Visual Data:} RGB images at a fixed resolution along with corresponding instance segmentation masks.
    \item \textbf{Object Annotations:} Visible objects are aligned with simulator metadata via unique identifiers. Category labels are normalized (lowercased with special characters removed), and compact 2D bounding boxes are computed from instance masks, filtering out any degenerate cases.
    \item \textbf{Relational Annotations:} OvSGTR/VG-style $\langle$subject, predicate, object$\rangle$ triples are generated to capture inter-object dependencies.
\end{itemize}

To support \textit{geometry-aware} relations, our generator integrates image-space topology, depth ordering, and 3D metric geometry. The relational hierarchy includes an image-topology layer (e.g., \textit{left of, right of, in front of, behind}) based on pixel-space proximity and median depth, and a geometric layer (e.g., \textit{on, above, below, in}) determined by 3D physical overlap, vertical displacement, and containment links provided by the simulator.

This multi-scene, multi-view synthesis strategy, by combining complementary image-plane signals with metric 3D data, significantly enhances the robustness and generalization of the scene graph model against viewpoint variations, object articulations, and occlusions.

\subsubsection{SceneDiver Adapter}
Our SceneDiver Adapter implements structured latent space reasoning through two core modules: Task-Guided Slot Attention and Scene-Aware MaskNet. The adapter is integrated between the visual projector and the LLM backbone of the VLA, operating directly within the vision-language alignment space.

\textbf{Task-Guided Slot Attention.}
This module decomposes projected visual tokens (dimension $D=4096$) into $K$ object-centric slot representations (dimension $d_s=256$). Departing from the random initialization of standard Slot Attention, we utilize a Task-Conditional Initialization strategy. First, a global task vector $v_{task}$ is extracted from the task token sequence using a learned attention pooling mechanism:
\begin{equation}
w_i = \frac{\exp(s_i)}{\sum_j \exp(s_j)}, \quad v_{task} = \sum_i w_i \cdot h_i
\end{equation}
where $s_i = \text{MLP}(t_i)$ represents the attention score for the $i$-th task token. The task vector is then projected to the slot dimension to serve as the mean for sampling, combined with learnable slot-specific offsets $\delta_k$ and a global variance $\sigma$:\begin{equation}S_0^k \sim \mathcal{N}(\text{Proj}(v_{task}) + \delta_k, \sigma)\end{equation} Slots are refined through $T=5$ iterations. In each iteration, we compute the slot-patch attention:\begin{equation}A_{kl} = \frac{\exp(\langle Q_k, K_l \rangle / \sqrt{d_s} \cdot \tau)}{\sum_{k'} \exp(\langle Q_{k'}, K_l \rangle / \sqrt{d_s} \cdot \tau)}\end{equation} where $\tau=0.4$ is a temperature parameter (below 1.0 to sharpen slot assignments and separate small objects). The weighted sum is fed into a GRUCell for state updates, followed by an MLP refinement. 

\textbf{Scene-Aware MaskNet.}
This module predicts the final focus mask based on refined slots using a coarse-to-fine approach:\paragraph{Slot-level Relevance Scoring (Coarse).} For each slot $s_k$, we fuse three signals: (i) slot semantic features $s_k$ (identity); (ii) \textbf{slot mass} $m_k = \sum_l A_{kl} / L$, projected via a 1$\to$32D MLP to reflect scene coverage; and (iii) the global task context $v_{task}$. These are concatenated to predict a relevance logit $r_k \in \mathbb{R}$:\begin{equation}r_k = \text{MLP}([\text{LN}(s_k); \text{Proj}(m_k); \text{Proj}(v_{task})])\end{equation}\paragraph{Patch-level Residual Refinement (Fine).} Slot semantics are back-propagated to the patch level using the attention map $A$ to build hybrid features $f_l = \sum_k A_{kl} \cdot s_k$. Combined with task context, an independent MLP predicts a local correction signal $\Delta_l$. The final patch logits are:\begin{equation}M_l = \sum_k r_k \cdot A_{kl} + \alpha \cdot \Delta_l\end{equation}where $\alpha$ is initialized near 0 to ensure reliance on the slot-level baseline early in training. A learnable scale parameter $\exp(\beta)$ (initially $\beta=1.1 \Rightarrow \text{scale}\approx 3$) is applied to slot logits to prevent ambiguous sigmoid outputs near 0.5.

\subsubsection{Experiment Details}
\textbf{Robot Manipulation.}
To increase task difficulty, we introduce visual distractors in the scene. Specifically, we generate two types of distractor objects: (1) a fake assembly board with random grooves on its surface, placed in a separate region from the real task board, and (2) 3-5 small cubes with random sizes (6-10 voxels per side) and colors, scattered in off-board areas. These distractors share similar visual features with task-relevant objects but are not part of the assembly goal, requiring the model to distinguish between relevant and irrelevant objects.

\textbf{Room Navigation.}
To select challenging target objects, we employ a distance-based sampling strategy using AI2-THOR scenes. For each floorplan, we filter small objects (volume $< 0.05$ m$^3$) that are visible and not structural elements (e.g., walls, floors). From these candidates, we either select the farthest object from the agent's initial position or randomly sample one that satisfies a minimum distance threshold. This ensures target objects are sufficiently challenging to locate and navigate to.

\textbf{Libero-plus.}
For the Libero-plus benchmark, we use the same scene generation and task setup as the original LIBERO benchmark without introducing additional modifications.

\subsubsection{Additional Qualitative Results}
\textbf{Room Navigation.} In this section, we present detailed results for the navigation task across its subtasks. Each subtask uses a distinct instruction to probe different facets of the model’s decision making. For every subtask, we provide two decision trajectories along with the corresponding instruction. For readability, we highlight the focused targets with red bounding boxes. Note that all annotations are invisible to the VLMs.

{\bf Base.} In this subtask, the instruction is very explicit, for example: “navigate to the pot in the room and get as close as possible to it.” The VLMs can directly observe the target object in the scene.
\begin{figure}[H]
\begin{center}
\includegraphics[width=0.9\linewidth]{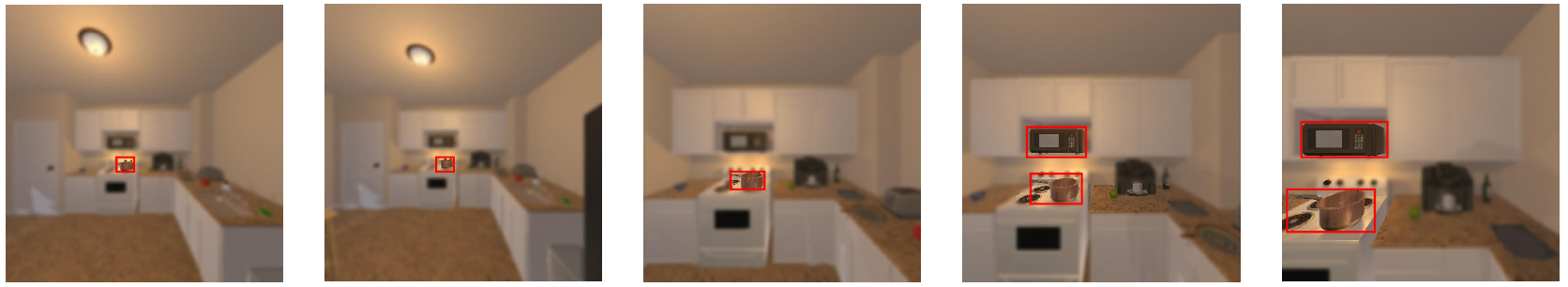}
\end{center}
\caption{Instruction: navigate to the Pot in the room and be as close as possible to it.}
\label{fig:navb1}
\end{figure}

\begin{figure}[H]
\begin{center}
\includegraphics[width=0.9\linewidth]{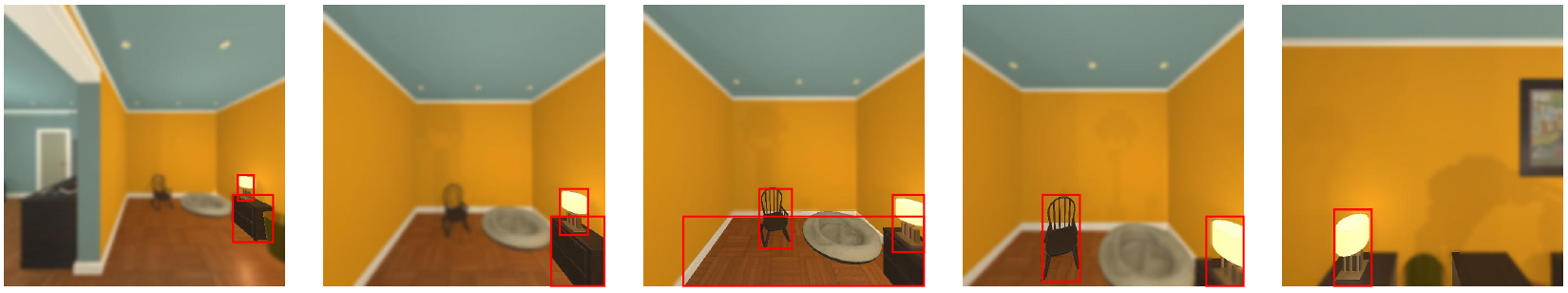}
\end{center}
\caption{Instruction: navigate to the DeskLamp in the room and be as close as possible to it.}
\label{fig:navb2}
\end{figure}
{\bf Common Sense.} Compared to the base task, this task involves a more complex instruction, requiring stronger reasoning capabilities from the MLLMs.
\begin{figure}[H]
\begin{center}
\includegraphics[width=0.9\linewidth]{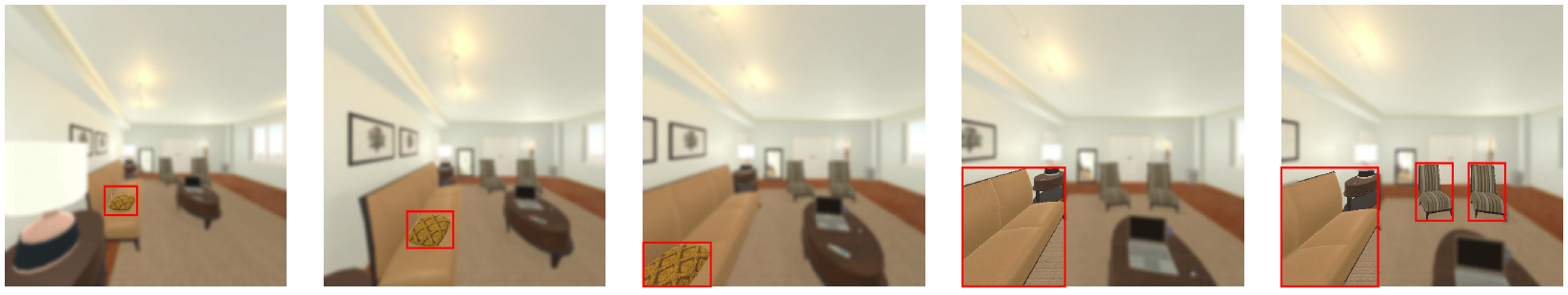}
\end{center}
\caption{Instruction: I need a soft cushion to support my head while sleeping. Can you navigate to that object and stay close?}
\label{fig:navc1}
\end{figure}

\begin{figure}[H]
\begin{center}
\includegraphics[width=0.9\linewidth]{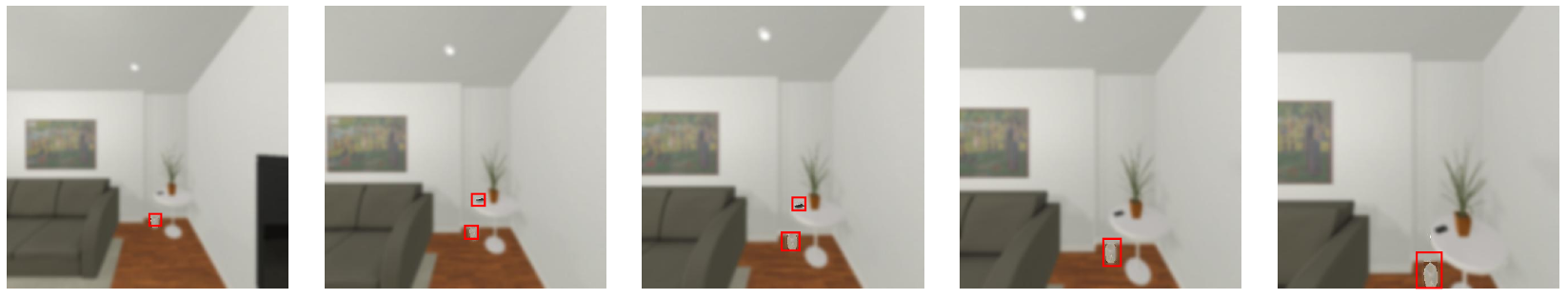}
\end{center}
\caption{Instruction: I'd like to view a decorative sculpture representing a figure or person. Can you navigate to that object and stay close?}
\label{fig:navc2}
\end{figure}
{\bf Complex Instruction.}
In this task, the instructions often contain substantial content irrelevant to the target, requiring VLMs to identify the task objective within complex prompts and align it with the objects in the image.
\begin{figure}[H]
\begin{center}
\includegraphics[width=0.9\linewidth]{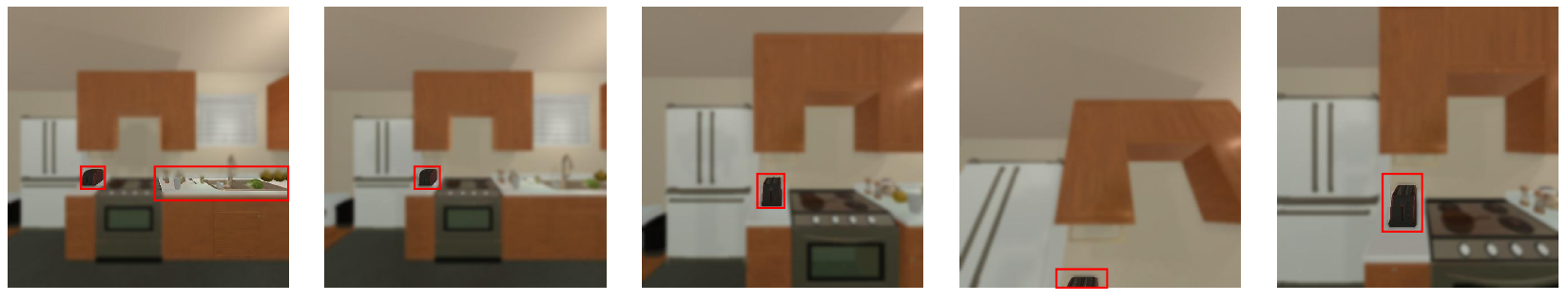}
\end{center}
\caption{Instruction: The sound of someone walking upstairs adds a subtle rhythm to the quiet morning. There's a folded towel on the counter, and the air smells faintly of butter. Could you navigate to the toaster for me? It's a peaceful start to the day.}
\label{fig:navci1}
\end{figure}

\begin{figure}[H]
\begin{center}
\includegraphics[width=0.9\linewidth]{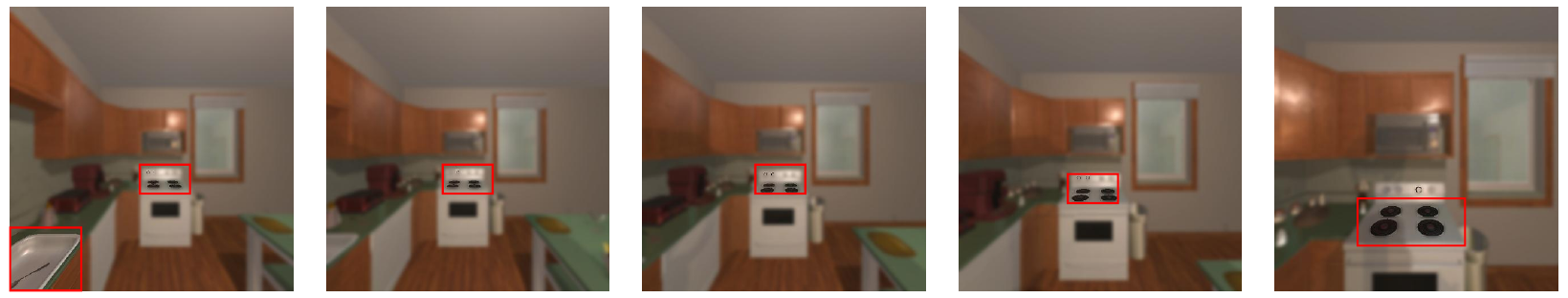}
\end{center}
\caption{Instruction: The rhythmic ticking of the kitchen clock blends with the occasional drip from the faucet. There's a small pile of onions on the table, freshly chopped. Please move towards the stove burner for me. The kitchen has a comforting hum to it.}

\label{fig:navci2}
\end{figure}
{\bf Visual Appearance Task.} The instruction in this task describes the visual appearance of the target object, requiring the MLLMs to possess strong visual comprehension capabilities.
\begin{figure}[H]
\begin{center}
\includegraphics[width=0.9\linewidth]{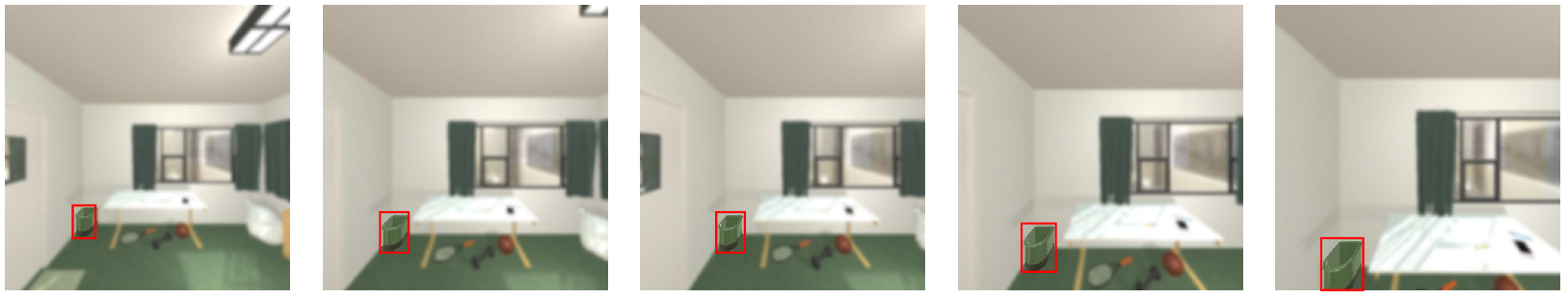}
\end{center}
\caption{Instruction: Approach the tall green container with a smooth texture.}
\label{fig:navv1}
\end{figure}

\begin{figure}[H]
\begin{center}
\includegraphics[width=0.9\linewidth]{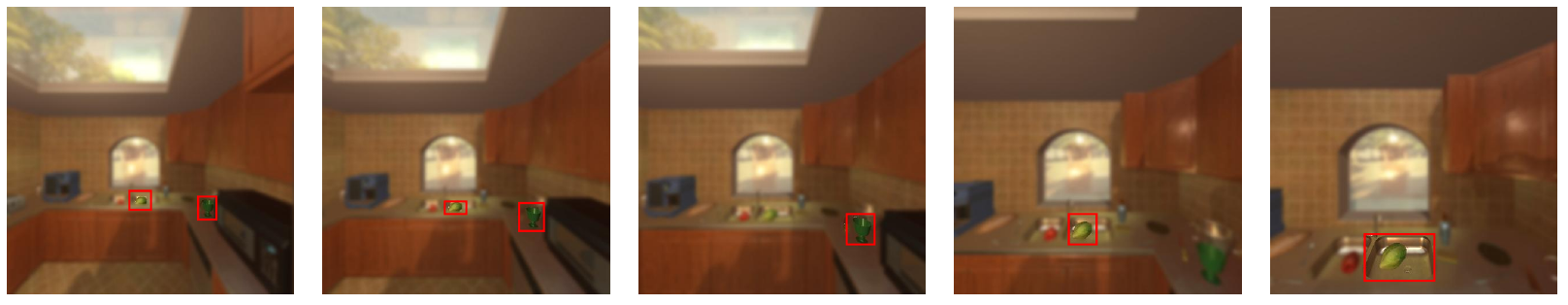}
\end{center}
\caption{Instruction: Move closer to the small round object with a green surface and a cylindrical shape.}
\label{fig:navv2}
\end{figure}

\textbf{Libero-plus.} In this section, we present the qualitative results of the SceneDiver Adapter on the LIBERO benchmark, illustrating the masking outcomes across various scenarios. Our method does not merely memorize objects from the training data; instead, it inherently learns to identify task-relevant entities and anticipate potential interactions during execution. This demonstrates that the adapter has genuinely mastered the ability to focus on critical visual cues. Furthermore, our approach maintains a degree of generalization even in unseen configurations, such as camera shifts or robotic arm displacements.

\begin{figure}[H]
    \centering
    \includegraphics[width=0.8\textwidth]{"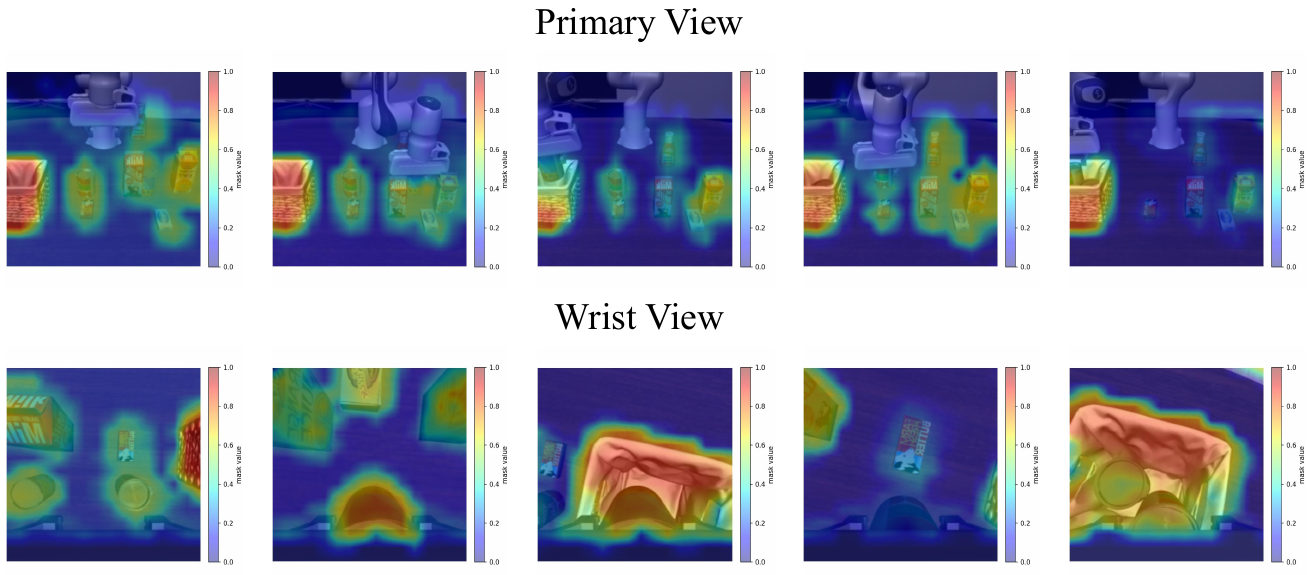"}
    \caption{Instruction: Put both the alphabet soup and the tomato sauce in the basket.}
\end{figure}

\begin{figure}[H]
    \centering
    \includegraphics[width=0.8\textwidth]{"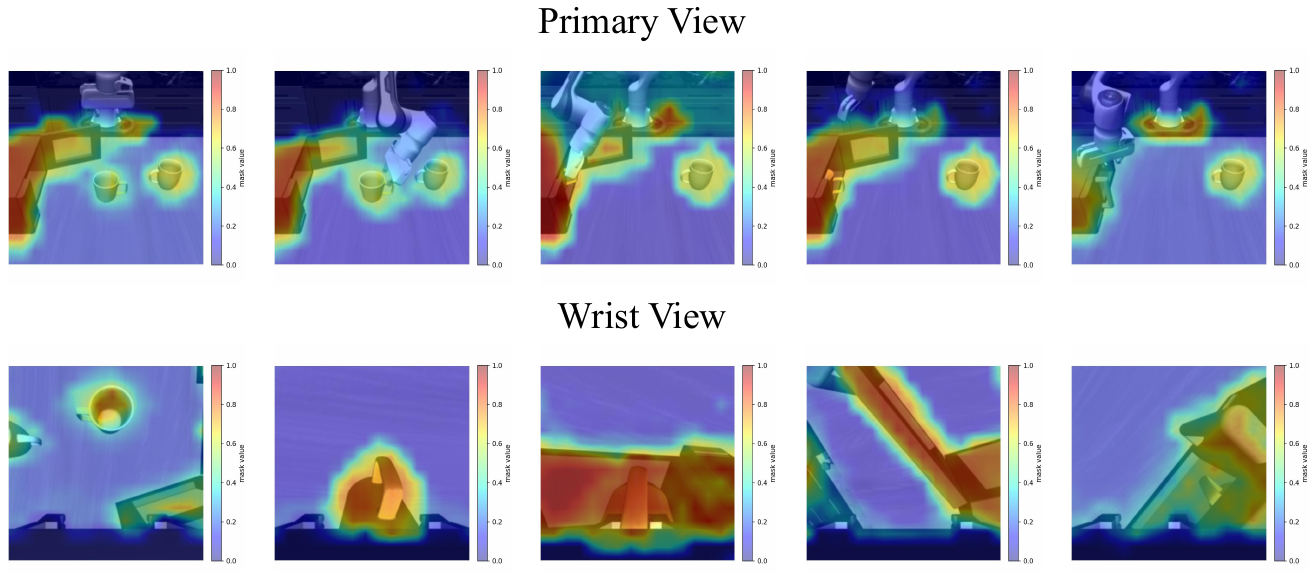"}
    \caption{Instruction: Put the yellow and white mug in the microwave and close it.}
\end{figure}

\begin{figure}[H]
    \centering
    \includegraphics[width=0.8\textwidth]{"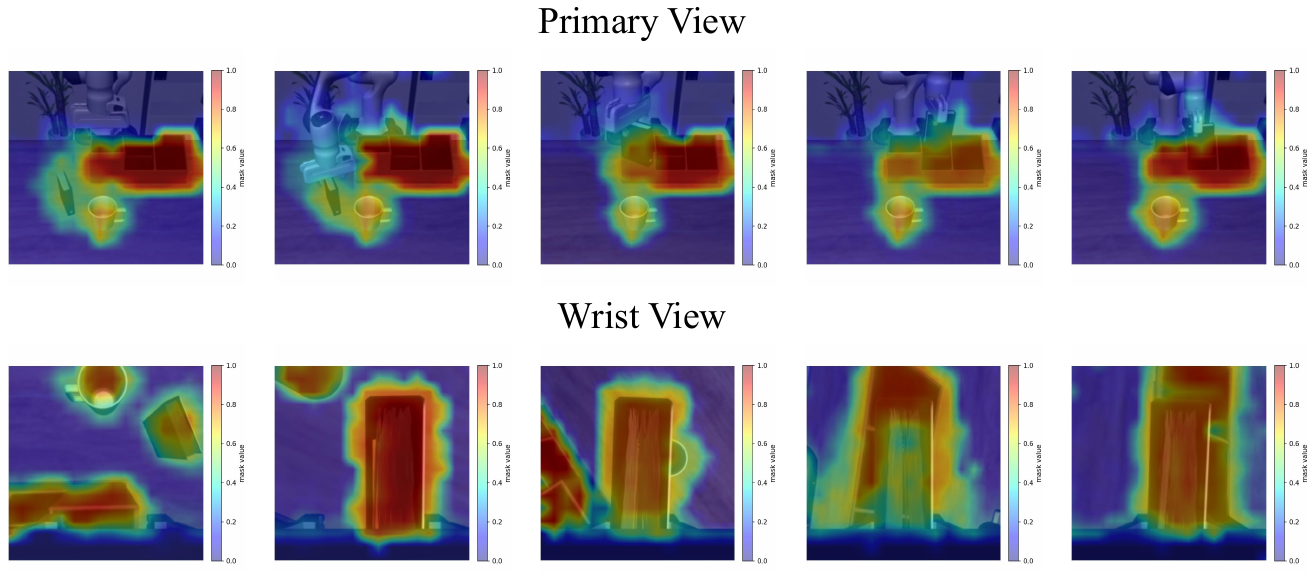"}
    \caption{Instruction: Pick up the book and place it in the back compartment of the caddy.}
\end{figure}

\begin{figure}[H]
    \centering
    \includegraphics[width=0.8\textwidth]{"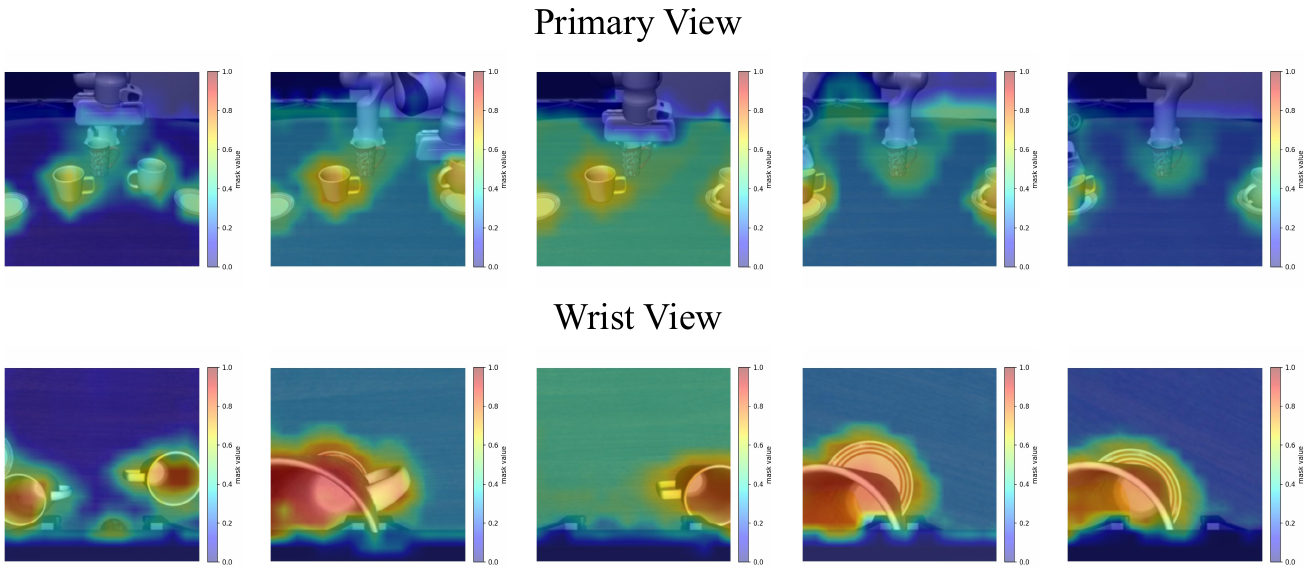"}
    \caption{Instruction: Put the white mug on the left plate and put the yellow and white mug on the right plate.}
\end{figure}

\begin{figure}[H]
    \centering
    \includegraphics[width=0.8\textwidth]{"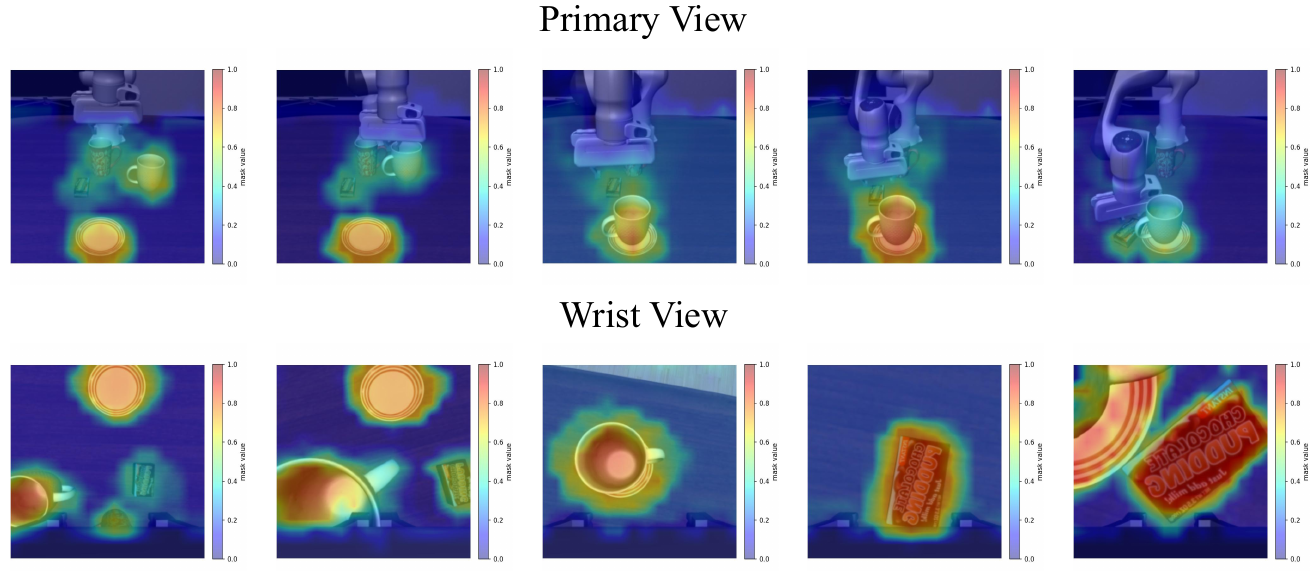"}
    \caption{Instruction: Put the white mug on the plate and put the chocolate pudding to the right of the plate.}
\end{figure}


\end{document}